\documentclass[10pt,twocolumn,letterpaper]{article}

\usepackage[pagenumbers]{cvpr} 

\usepackage{algorithmic}
\usepackage{algorithm}
\usepackage{gensymb}

\usepackage{capt-of}
\usepackage{caption}
\usepackage{booktabs}
\usepackage{multirow}
\usepackage{makecell}
\usepackage{graphicx}
\usepackage{array}
\usepackage{tabularx}
\usepackage[table]{xcolor}
\definecolor{lightpurblue}{RGB}{243, 249, 255}
\definecolor{midpurblue}{RGB}{205, 227, 255}
\definecolor{lightgreen}{RGB}{235, 255, 240}
\definecolor{midgreen}{RGB}{233, 253, 237}
\definecolor{lightred}{RGB}{254, 240, 236}
\newcommand{\best}[1]{\cellcolor{midpurblue}{#1}}
\newcommand{\secondbest}[1]{\cellcolor{lightpurblue}{#1}}

\definecolor{cvprblue}{rgb}{0.21,0.49,0.74}
\usepackage[pagebackref,breaklinks,colorlinks,allcolors=cvprblue]{hyperref}

\def\confName{CVPR}
\def\confYear{2027}

\title{EchoACE: Action-Centric Embodied Echocardiography Foundation Model}

\author{
Teng Wang$^{1,*}$, 
Haojun Jiang$^{2,*,\ddagger}$, 
Jiaqi Yu$^{3}$, 
Chenxi Li$^{1}$, 
Diwen Wang$^{4}$\\[0.3em]
Yang Yue$^{1}$, 
Yihang Tang$^{4}$, 
Gao Huang$^{1,5,\dagger}$\\[0.8em]
{\normalsize
$^{1}$ Department of Automation, Tsinghua University
\hspace{1em}
$^{2}$ Johns Hopkins University
}\\[0.1em]
{\normalsize
$^{3}$ Beijing University of Posts and Telecommunications
\hspace{1em}
$^{4}$ Beijing Institute of Technology
}\\[0.1em]
{\normalsize
$^{5}$ Institute for Embodied Intelligence and Robotics, Tsinghua University
}
}

\begin{document}
\maketitle

%%%%%
\begingroup
\renewcommand{\thefootnote}{}
\footnotetext{
\textsuperscript{*} Equal contribution.
\
\textsuperscript{$\ddagger$} Guided this work.
\
\textsuperscript{$\dagger$} Corresponding author.
}
\endgroup

\begin{abstract}
Embodied echocardiography, which aims to automate or assist with cardiac ultrasound scanning, requires multiple complementary capabilities: deciding probe movements, predicting their visual outcomes, inferring past probe motion, and evaluating scanning progress.
All these capabilities share a common foundation in understanding the physical displacements, \textit{i.e.}, probe actions, between ultrasound views.
However, existing foundation models are predominantly appearance-centric, organizing their latent spaces by visual similarity.
In echocardiography, observations from the same probe pose can differ substantially due to cardiac motion, respiration, and imaging settings, while those from distant poses can appear similar.
As a result, their latent spaces fail to preserve the physical proximity of ultrasound views.
In this paper, we present EchoACE, a foundation model that moves beyond appearance-centric pretraining toward action-centric representation learning.
The central idea is to reshape ultrasound representation geometry through a Physical Distance Ranking objective that encourages latent distances to preserve the ordering of physical displacements.
Pretrained on over one million echocardiography samples, EchoACE provides a shared representation that consistently outperforms the strongest baseline across four embodied echocardiography tasks, reducing errors by up to 46.2\% and raising the correlation between latent and physical distances from 0.43 to 0.93.
Remarkably, its action-centric pretraining stage reduces compute by 285$\times$ compared with the strongest baseline.
Code will be released upon acceptance.
\end{abstract}    
\section{Introduction}
\label{sec:intro}

Echocardiography is widely used to assess cardiovascular diseases owing to its real-time, radiation-free, and low-cost nature~\cite{roth2017global}.
In routine echocardiography, the probe captures only a single cross-section of the three-dimensional heart at a time. 
Sonographers must therefore continuously adjust the probe pose based on ultrasound feedback to visualize key chambers, vessels, and valves.
This expertise requires extensive training, which leads to workforce shortages~\cite{won2024sound}.
These challenges have motivated the emergence of embodied echocardiography, which aims to automate or assist with cardiac ultrasound scanning.
In the long term, this technology has the potential to broaden access to high-quality echocardiography worldwide~\cite{jiang2023robotic}.

\begin{figure}[!t]
    \centering
    \includegraphics[width=\linewidth]{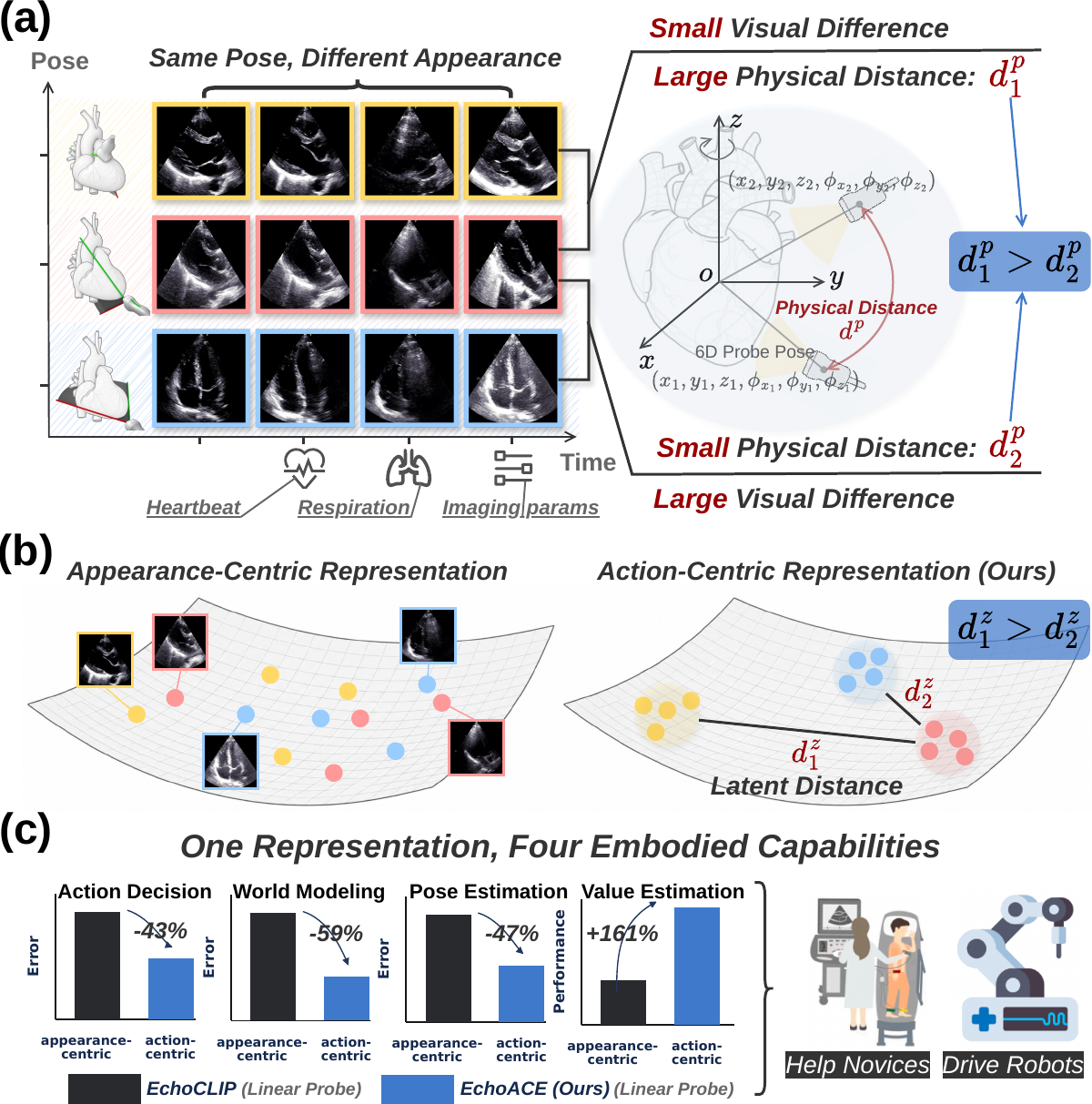}
    \vspace{-12pt}
    \caption{
        \textbf{Overview}. 
        (a) Mismatch between visual difference and physical distance.
        (b) Illustration of appearance-centric and action-centric representation spaces.
        (c) Superior performance on four representative embodied echocardiography tasks.
    }
    \label{fig:overview}
    \vspace{-13pt}
\end{figure}

\begin{figure*}[h]
    \centering
    \includegraphics[width=\linewidth]{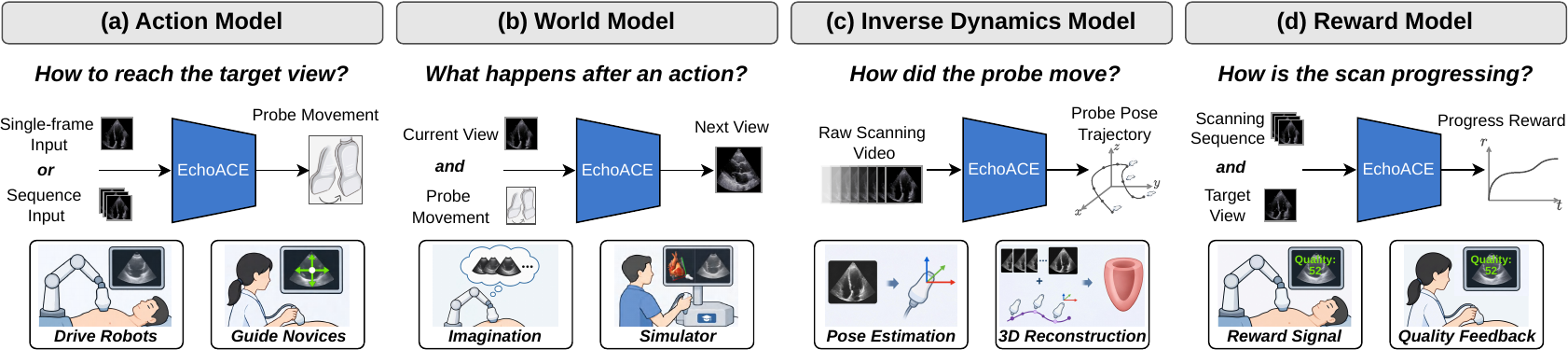}
    \vspace{-15pt}
    \caption{\textbf{Embodied echocardiography tasks}.
    (a) Action model for deciding probe movements.
    (b) World model for predicting future visual outcomes.
    (c) Inverse dynamics model for inferring past probe motion.
    (d) Reward model for evaluating scanning progress.
    }
    \label{fig:task}
    \vspace{-12pt}
\end{figure*}

Realizing this potential requires embodied echocardiography systems to master several complementary capabilities, including deciding probe movements, predicting their visual outcomes, inferring past probe motion, and evaluating scanning progress.
All these capabilities build upon understanding the physical displacements between ultrasound views.
This suggests that a single shared representation could serve all of them, provided that its latent space preserves the physical proximity between views.
However, existing foundation models~\cite{jiao2024usfm,mh2024lvm,zhang2023biomedclip,christensen2024vision,jiang2024structure,yue2025echoworld,jiang2025ultrasep} rarely preserve such physical proximity in their latent spaces.
A major line of work~\cite{jiao2024usfm,mh2024lvm,zhang2023biomedclip,christensen2024vision} learns an appearance-centric latent space, as it relies primarily on appearance-level pretraining supervision.
For example, USFM~\cite{jiao2024usfm} is pretrained via spatial-frequency dual masked image modeling, reconstructing masked content in both the spatial and frequency domains. 
Since these objectives are defined purely on image content, the learned representations are dominated by visual appearance.
In echocardiography, however, visual similarity does not imply physical proximity.
As shown in the first row of Fig.~\ref{fig:overview}(a), observations at the same probe pose can vary in appearance due to respiration, cardiac motion, and imaging parameters, which pushes them apart in the latent space (Fig.~\ref{fig:overview}(b) left).
Moreover, observations acquired at distant probe poses can share similar anatomical structures (first two rows) and are thus mapped close together.
Conversely, small probe displacements may expose different anatomical structures (last two rows), so physically adjacent observations can end up far apart in the latent space.
Such misalignment between latent and physical proximity hinders the transfer of these representations to embodied echocardiography tasks.

Another line of work~\cite{yue2025echoworld,jiang2025ultrasep,jiang2024structure} incorporates probe motion through action-conditioned pretraining. For example, EchoWorld~\cite{yue2025echoworld} and UltraSeP~\cite{jiang2025ultrasep} predict the resulting view's latent features conditioned on probe actions. However, supervision remains image-feature-based, with actions serving only as input conditions, thus weakly constraining latent-space structure. As a result, representations remain appearance-dominated and fail to preserve physical proximity. Consequently, neither line of work yields representations that faithfully reflect physical relationships, limiting their transfer to embodied echocardiography tasks.

In this paper, we propose EchoACE, an action-centric foundation model for embodied echocardiography, in which physical probe actions, rather than visual appearance alone, determine the geometry of the latent space. 
To this end, EchoACE first equips the model with basic echocardiography visual representations through self-supervised pretraining, and then uses physical probe displacements as direct supervision to further shape its latent geometry.
Specifically, we introduce a Physical Distance Ranking objective: given pairs of ultrasound observations with known relative probe poses, it encourages the ordering of their latent distances to be consistent with that of their physical displacements. 
This ordinal formulation avoids imposing an arbitrary scale on the unitless latent space and remains robust to inter-subject anatomical variability, under which absolute distances between views vary across patients while their relative proximity remains consistent.
Moreover, unlike action-conditioned methods that rely on an additional predictor~\cite{yue2025echoworld,jiang2025ultrasep,jiang2024structure}, our objective acts directly on the encoder, embedding physical structure into the representations used by downstream tasks.
Consequently, EchoACE can serve as a shared foundation for diverse embodied echocardiography capabilities.

We pretrain EchoACE on 1.31 million echocardiography images paired with their corresponding probe poses and transfer it to four representative tasks: probe navigation~\cite{jiang2025ultrasep}, action-conditioned world modeling~\cite{yue2025echoworld}, inverse dynamics prediction~\cite{iijima2026echopose}, and progress reward~\cite{narang2021utility} (Fig.~\ref{fig:task}). 
We compare EchoACE with both appearance-centric and action-conditioned baselines.
The appearance-centric baselines include general vision foundation models~\cite{tschannen2025siglip,simeoni2025dinov3}, classic self-supervised methods~\cite{mae,mocov3} pretrained on the same data, as well as open-source echocardiography~\cite{christensen2024vision}, ultrasound~\cite{jiao2024usfm}, and medical~\cite{zhang2023biomedclip,mh2024lvm} foundation models with about four times more encoder parameters.
The action-conditioned baselines~\cite{yue2025echoworld,jiang2025ultrasep,jiang2024structure} are pretrained on the same data.
To directly compare the learned representations, we adopt a frozen-encoder setting, in which all methods share identical task-specific heads. 
Under this setting, EchoACE achieves the best performance on all metrics across the four tasks. 
Compared with the strongest baselines, it reduces errors by up to 30.8\% in probe navigation, 46.2\% in world modeling, and 45.8\% in inverse dynamics prediction. 
For progress reward, it raises the Spearman correlation between latent and physical distances from 0.43 to 0.93.
Remarkably, these gains come at a negligible cost: sharing the same first-stage pretraining as UltraSeP, EchoACE’s second stage takes only 0.6 GPU-hours versus 171.0 GPU-hours, a 285$\times$ reduction.

\section{Related Work}
\noindent\textbf{Embodied Echocardiography.}
This field encompasses tasks that require reasoning about the interplay between probe actions and ultrasound observations, supporting both AI-assisted scanning by clinicians and autonomous robotic scanning.
Probe navigation, the most extensively studied task, is typically learned from expert demonstrations~\cite{narang2021utility,bao2024real,jiang2025ultrasep,wang2026ultrastar,wang2026va,wang2024robotic,jiang2024intelligent,chen2025uspilot,jiang2025towards,wang2026ultrahit} or through reinforcement learning in simulated or robotic environments~\cite{amadou2024goal,shida2025robotic,zakeri2025deep,cao2026anatomical}.
Beyond deciding where to move, world models~\cite{yue2025echoworld,jiang2024structure} predict the latent features of the view reached after a probe movement, primarily serving as pretraining objectives for navigation.
Conversely, inverse dynamics infers probe motion from ultrasound observations~\cite{yu2025ultratwin,laumer20252d}, enabling sensorless freehand 3D reconstruction~\cite{luo2023recon} and 6D pose estimation of sparse echocardiograms for left-ventricular shape reconstruction~\cite{iijima2026echopose}.
To assess scanning progress, existing systems typically rely on image quality~\cite{narang2021utility,zakeri2025deep} or anatomical matching~\cite{cao2026anatomical}, which reflect visual appearance rather than physical proximity to target view.
These methods mostly target individual capabilities, whereas EchoACE serves as a single shared foundation model that supports all of them.

\noindent \textbf{Ultrasound Foundation Models.}
These models learn transferable representations through large-scale pretraining.
Most existing ones are appearance-centric, with objectives defined on image content or its textual descriptions: masked image modeling reconstructs masked content in the pixel, frequency, or latent space~\cite{mae,jiao2024usfm,ijepa}, contrastive and self-distillation methods enforce invariance across augmented views~\cite{mocov3,mh2024lvm,oquab2023dinov2}, and vision-language pretraining aligns images with captions or clinical reports~\cite{zhang2023biomedclip,christensen2024vision,guo2026visually}.
Consequently, their latent spaces are organized by visual appearance and semantics, such that physically distant views with similar anatomical structures may lie close together.
To incorporate probe motion, recent echocardiography pretraining methods~\cite{jiang2024structure,jiang2025ultrasep,yue2025echoworld} predict the features of the resulting view conditioned on probe actions or jointly model image--action sequences.
However, their supervision still comes primarily from image features, leaving the encoder's latent geometry unconstrained by physical displacements.
In contrast, EchoACE directly structures the encoder’s latent space by physical probe displacements, yielding representations whose distances reflect physical proximity and can be shared across embodied tasks.

\section{Embodied Echocardiography Tasks}
\label{sec:tasks}

Embodied echocardiography requires several complementary capabilities, including predicting probe movements, anticipating their visual outcomes, inferring past probe motion, and evaluating scanning progress. We formulate four tasks to assess these capabilities, as illustrated in \cref{fig:task}.

\begin{figure*}[!t]
    \centering

    % Left: Dataset
    \begin{minipage}[t]{0.30\linewidth}
        \vspace{0pt}
        \centering
        \includegraphics[width=\linewidth]{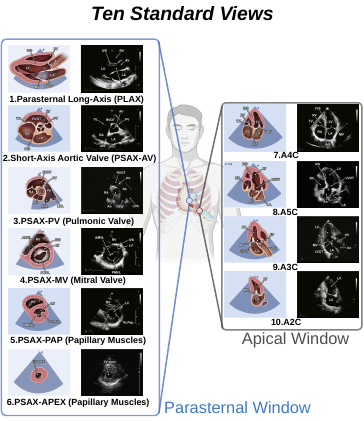}
        \caption{The ten standard cardiac views (images are adapted from ~\cite{mitchell2019guidelines}).}
        \label{fig:dataset}
    \end{minipage}
    \hfill
    % Right: Method
    \begin{minipage}[t]{0.68\linewidth}
        \vspace{0pt}
        \centering
        \includegraphics[width=\linewidth]{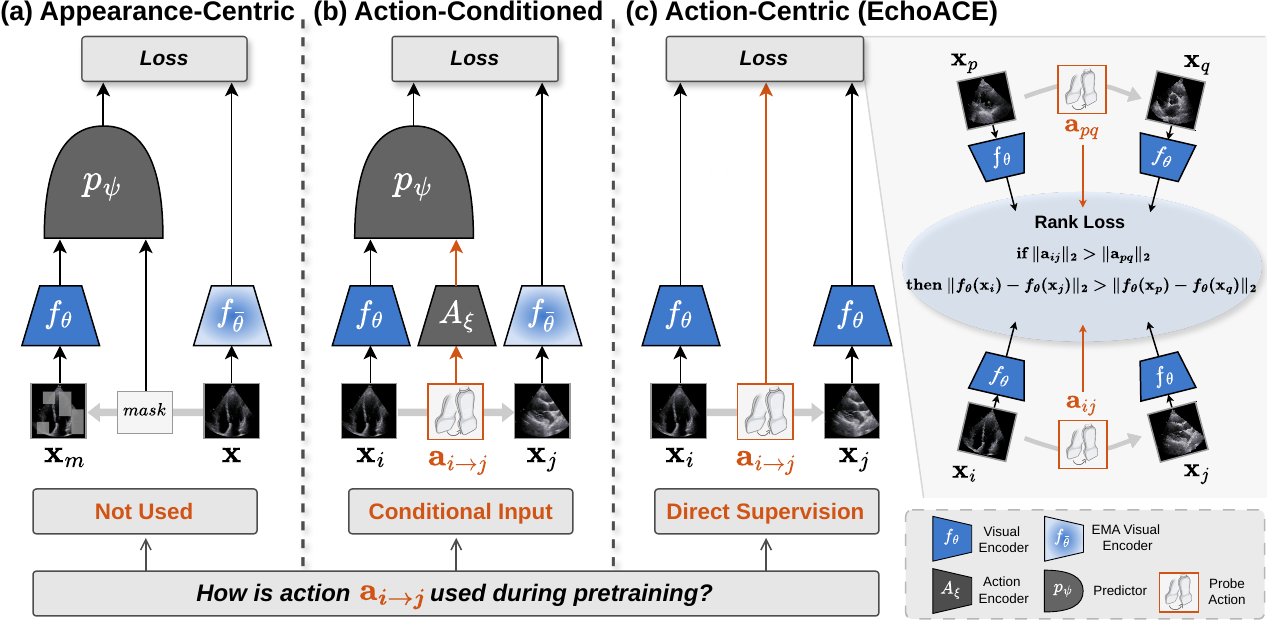}
        \caption{
            \textbf{Three pretraining paradigms.}
            (a) “Appearance-centric” do not incorporate action information.
            (b) “Action-conditioned” use actions only to condition visual prediction.
            (c) EchoACE directly structures the latent space using physical action, without an extra predictor.
        }
        \label{fig:method}
    \end{minipage}

    \vspace{-10pt}
\end{figure*}

\noindent\textbf{Action Prediction for Probe Navigation.}
Action prediction aims to estimate the probe movement required to reach a target standard cardiac view. It can guide novice sonographers in adjusting their probes and enable robotic systems to autonomously acquire standard cardiac views. Given an input observation $\mathbf{x}_t$, which can be a single ultrasound image or a history of images and probe actions, the model predicts the relative action toward the ten standard views (Fig.~\ref{fig:dataset}):
\begin{equation*}
\mathbf{z}_t=f_\theta(\mathbf{x}_t),\quad \hat{\mathbf{a}}_{t\rightarrow k}=h_\phi(\mathbf{z}_t,k),\quad k\in\{1,\ldots,10\}.
\end{equation*}
Here, $h_\phi$ denotes the action prediction head, $\hat{\mathbf{a}}_{t\rightarrow k}$ represents the predicted 6D relative probe movement, and $k$ indexes the target standard view.

\noindent\textbf{Action-Conditioned World Modeling.}
This task aims to predict the visual outcomes of probe movements. It can provide novice sonographers with an interactive ultrasound simulation environment for training and enable robotic systems to perform test-time imagination before executing actions. Given the current observation and probe action, the world model predicts the future representation, which can subsequently be decoded into an ultrasound image:
\begin{equation*}
\mathbf{z}_t=f_\theta(\mathbf{x}_t),\quad \hat{\mathbf{z}}_{t+1}=P_\psi(\mathbf{z}_t,\mathbf{a}_t),\quad \hat{\mathbf{x}}_{t+1}=D_\omega(\hat{\mathbf{z}}_{t+1}).
\end{equation*}
Here, $P_\psi$ denotes the action-conditioned world model predictor, and $D_\omega$ denotes the image decoder.

\noindent\textbf{Inverse Dynamics Prediction.}
This task aims to infer the relative probe movement between two observations. It can facilitate trackerless probe pose estimation and trackerless 3D cardiac reconstruction. Given two observations $\mathbf{x}_i$ and $\mathbf{x}_j$, the model predicts their relative probe movement:
\begin{equation*}
\mathbf{z}_i=f_\theta(\mathbf{x}_i),\quad \mathbf{z}_j=f_\theta(\mathbf{x}_j),\quad \hat{\mathbf{a}}_{i\rightarrow j}=g_\eta(\mathbf{z}_i,\mathbf{z}_j).
\end{equation*}
Here, $g_\eta$ denotes the inverse dynamics prediction head.

\noindent\textbf{Latent Distance as a Progress Reward.} 
Progress estimation evaluates how close the current probe pose is to a desired target view. 
It can provide real-time feedback to novice sonographers during probe adjustment and serve as a reward signal for autonomous robotic scanning. 
Given a current observation $\mathbf{x}_t$ and a target observation $\mathbf{x}_g$, the latent distance between them is computed as:
\begin{equation*}
\mathbf{z}_t=f_\theta(\mathbf{x}_t),\quad \mathbf{z}_g=f_\theta(\mathbf{x}_g),\quad r_t^{\mathrm{dist}}=\|\mathbf{z}_t-\mathbf{z}_g\|_2.
\end{equation*}
If the latent distance reliably reflects the corresponding physical distance, it can serve directly as a progress signal without requiring an additional prediction network.

In the next section, we introduce EchoACE, which learns a unified representation $\mathbf{z}=f_\theta(\mathbf{x})$ shared by all four tasks.

\section{Method}
\label{sec:method}

The four embodied echocardiography tasks require representations that capture physical displacements while remaining robust to appearance variations. 
To this end, we propose EchoACE, an action-centric foundation model with a two-stage pretraining pipeline. Stage 1 learns a visual prior of cardiac ultrasound to initialize Stage 2, which explicitly structures the latent space of the shared encoder $\mathbf{z}=f_\theta(\mathbf{x})$ according to physical displacements through a Physical Distance Ranking objective, as illustrated in Fig.~\ref{fig:method}(c).

\subsection{Dataset}
\label{sec:dataset}

Our dataset comprises 178 subjects and 356 expert scanning trajectories, totaling 1.31 million synchronized ultrasound image--pose samples. 
Data collection was conducted with the approval and oversight of the University Medical Ethics Committee. 
Two senior sonographers with over 10 years of experience performed continuous scans of ten standard cardiac views (Fig.~\ref{fig:dataset}) using an ultrasound probe mounted on a robotic arm. During acquisition, the system synchronously recorded ultrasound images and the corresponding six-dimensional probe poses, while the sonographers annotated the ten standard views (Fig.~\ref{fig:dataset}). Each observation is represented as $(\mathbf{x}_t,\mathbf{T}_t)$, where $\mathbf{x}_t\in\mathbb{R}^{3\times H\times W}$ denotes the ultrasound image and $\mathbf{T}_t\in SE(3)$ denotes its corresponding probe pose. For any two observations, their relative probe transformation is obtained as $\Delta\mathbf{T}_{i\rightarrow j}=\mathbf{T}_i^{-1}\mathbf{T}_j$. These pose annotations provide the physical supervision for learning action-centric representations.

\subsection{Stage 1: Visual Prior Pretraining}

We first establish a visual prior of cardiac structures by pretraining the image encoder on 2D ultrasound images following the I-JEPA-style paradigm~\cite{ijepa}. This stage aims to capture general cardiac structural patterns from visual observations, providing a foundation for subsequent physical geometry modeling. Given an input image $\mathbf{x}$, we randomly sample visible context patches and multiple target blocks. The context encoder $f_\theta$ extracts features from the visible patches $\mathbf{x}_{C}$, while an exponential moving average (EMA) target encoder $f_{\bar{\theta}}$ provides target features from the full image. A lightweight predictor $p_\psi$ takes the context features and positional mask tokens $\mathbf{M}_{B_m}$ to predict the features of each target block $B_m$:
\begin{equation*}
\hat{\mathbf{F}}_{B_m}=p_\psi(f_\theta(\mathbf{x}_{C}),\mathbf{M}_{B_m}),\quad
\mathbf{F}_{B_m}=f_{\bar{\theta}}(\mathbf{x})|_{B_m}.
\end{equation*}
The encoder is trained to predict target representations rather than reconstruct image pixels, using the objective:
\begin{equation*}
\mathcal{L}_{\mathrm{2D}}
=
\frac{1}{M}
\sum_{m=1}^{M}
\left\|
\hat{\mathbf{F}}_{B_m}
-
\operatorname{sg}(\mathbf{F}_{B_m})
\right\|_2^2,
\end{equation*}
where $M$ is the number of target blocks and $\operatorname{sg}(\cdot)$ denotes stop-gradient. The context encoder and predictor are optimized by gradient descent, while the target encoder is updated as an EMA of the context encoder, i.e., $\bar{\theta}\leftarrow\mu\bar{\theta}+(1-\mu)\theta$, where $\mu$ is the EMA decay factor. After this stage, we retain the pretrained context encoder to initialize Stage 2, where physical probe displacements are introduced to explicitly structure the latent geometry.

\subsection{Stage 2: Action-Centric Pretraining}
Building on the visual prior from Stage 1, we leverage the relative probe poses between observations to explicitly structure the encoder’s latent space according to their physical proximity.
Given two ultrasound observations, we first compute their relative probe transformation:
\begin{equation*}
\Delta\mathbf{T}_{i\rightarrow j}
=
\mathbf{T}_i^{-1}\mathbf{T}_j
=
\begin{bmatrix}
\Delta\mathbf{R}_{ij} & \Delta\mathbf{t}_{ij}\\
\mathbf{0}^{\top} & 1
\end{bmatrix}.
\end{equation*}
Here, $\Delta\mathbf{R}_{ij}\in SO(3)$ and $\Delta\mathbf{t}_{ij}\in\mathbb{R}^3$ denote the relative rotation and translation, respectively. We represent the rotation using its axis-angle vector $\boldsymbol{\phi}_{ij}=(\operatorname{Log}(\Delta\mathbf{R}_{ij}))^\vee$, where $\operatorname{Log}$ denotes the rotation matrix logarithm and $(\cdot)^\vee$ converts a skew-symmetric matrix into a rotation vector.

Since translation and rotation have different physical units, we introduce a reference length $L$ (defaulting to $100\,\mathrm{mm}$ based on typical probe-motion scales, see Supplementary D.1 for a sensitivity analysis of this value), serving as an effective lever arm from the probe's rotation center. As $\boldsymbol{\phi}_{ij}$ is expressed in radians, $L\|\boldsymbol{\phi}_{ij}\|_2$ corresponds to the arc length swept by a point at distance $L$ under the relative rotation. We thus convert rotational displacement into an equivalent linear displacement and define the physical displacement vector $\tilde{\mathbf{a}}_{ij}$ and physical distance $d_{ij}^{p}$ as:
\begin{equation*}
\tilde{\mathbf{a}}_{ij}
=
\begin{bmatrix}
\Delta\mathbf{t}_{ij}\\
L\boldsymbol{\phi}_{ij}
\end{bmatrix},
\quad
d_{ij}^{p}
=
\|\tilde{\mathbf{a}}_{ij}\|_2
=
\sqrt{
\|\Delta\mathbf{t}_{ij}\|_2^2
+
L^2\|\boldsymbol{\phi}_{ij}\|_2^2
}.
\end{equation*}

For the corresponding ultrasound observations, the encoder $f_\theta$, initialized from Stage 1, extracts their global representations, and the latent distance $d_{ij}^{z}$ is computed as:
\begin{equation*}
\mathbf{z}_i=f_\theta(\mathbf{x}_i),\quad
\mathbf{z}_j=f_\theta(\mathbf{x}_j),\quad
d_{ij}^{z}=\|\mathbf{z}_i-\mathbf{z}_j\|_2.
\end{equation*}
The physical distance $d_{ij}^{p}$ measures probe displacement, whereas the latent distance $d_{ij}^{z}$ measures the separation between the corresponding representations.

Our key idea is to align the ordering of latent distances with that of physical distances.
Due to substantial inter-subject anatomical variability, the numerical scale of physical displacements can vary considerably across patients, making it difficult to establish a consistent mapping from physical distances to latent distances. 
In contrast, the relative ordering of physical proximities is more consistent across subjects. We therefore adopt an ordinal formulation that encourages latent distances to preserve the ranking of physical displacements.

Given a mini-batch of $B$ sampled image pairs, we denote their physical and latent distances as $d^p$ and $d^z$, respectively. For any two sampled pairs $q$ and $r$, if $d_q^p>d_r^p$, their latent distances should satisfy $d_q^z>d_r^z$. We formulate this constraint using a RankNet-style~\cite{burges2005learning} pairwise ranking objective over unordered comparisons with distinct physical distances: $\mathcal{Q}=\{(q,r):1\leq q<r\leq B,\ d_q^p\neq d_r^p\}$. For each valid comparison, the ranking label is defined as $s_{qr}=\operatorname{sign}(d_q^p-d_r^p)$. The resulting ranking loss is computed by averaging over all valid comparisons:
\begin{equation*}
\mathcal{L}_{\mathrm{rank}}
=
\frac{1}{|\mathcal{Q}|}
\sum_{(q,r)\in\mathcal{Q}}
\operatorname{softplus}
\left(
-\frac{s_{qr}(d_q^z-d_r^z)}{\tau}
\right).
\end{equation*}
Here, $\tau$ is the ranking temperature, set to $0.1$ by default. Starting from the Stage 1 initialization, we optimize the encoder directly using this Physical Distance Ranking objective. 
The Stage 1 predictor and target encoder are no longer required, making Stage 2 a simple yet effective way to embed physical probe geometry into the shared representations without additional predictive networks.

\section{Experiments}
\label{sec:experiments}

% Method and Encoder columns
\newcolumntype{Q}{
    >{\raggedright\arraybackslash}m{2.40cm}
}

\newcolumntype{R}{
    >{\centering\arraybackslash}m{1.05cm}
}

% Action Model + World Model: 9 columns
\newcolumntype{W}{
    >{
        \hsize=0.95\hsize
        \linewidth=\hsize
        \centering\arraybackslash
    }X
}

% Inverse Dynamics: 2 wider columns
\newcolumntype{I}{
    >{
        \hsize=1.15\hsize
        \linewidth=\hsize
        \centering\arraybackslash
    }X
}

% Reward Model: 3 columns
\newcolumntype{S}{
    >{
        \hsize=1.05\hsize
        \linewidth=\hsize
        \centering\arraybackslash
    }X
}

\newcolumntype{K}{
    >{
        \hsize=1.05\hsize
        \linewidth=\hsize
        \centering\arraybackslash
    }X
}

\newcolumntype{P}{
    >{
        \hsize=1.05\hsize
        \linewidth=\hsize
        \centering\arraybackslash
    }X
}

% Unified thin gray vertical rule
\providecommand{\grayvrule}{}
\renewcommand{\grayvrule}{%
    \color{gray!65}\vrule width 0.35pt
}

\providecommand{\gain}[1]{}

\renewcommand{\gain}[1]{%
    {\color{gray!70}%
    \fontsize{7}{8.5}\selectfont
    \strut #1}%
}

\begin{table*}[t]
    \centering

    \caption{
    \textbf{Main results.}
    \textbf{(1) The Action Model} is evaluated under single-frame and sequence-input protocols using translation/rotation mean absolute error (T-MAE/R-MAE).
    \textbf{(2) The World Model} is evaluated using T-MAE and R-MAE for latent retrieval, together with Learned Perceptual Image Patch Similarity (LPIPS), Fr\'echet Inception Distance (FID), and Structural Similarity Index Measure (SSIM) for image generation.
    \textbf{(3) The Inverse Dynamics Model} is evaluated using T-MAE and R-MAE.
    \textbf{(4) The Reward Model} is evaluated using Spearman's rank correlation coefficient ($\rho$), Kendall's rank correlation coefficient ($\tau$), and pairwise ranking accuracy.
    Translation and rotation errors are measured in millimeters and degrees, respectively.
    \dag~indicates that use only their respective pre-training strategies, while keeping the downstream heads identical for a fair comparison.
    }

    \label{tab:result_main}

    \vspace{-2mm}

    \begingroup

    \footnotesize

    \setlength{\tabcolsep}{0.8pt}

    \renewcommand{\arraystretch}{1.050}

    \renewcommand{\tabularxcolumn}[1]{m{#1}}

    \begin{tabularx}{\textwidth}{
        @{}
        Q R
        !{\grayvrule}
        W W
        !{\grayvrule}
        W W
        !{\grayvrule}
        W W
        !{\grayvrule}
        W W W
        !{\grayvrule}
        I I
        !{\grayvrule}
        S K P
        @{}
    }

        \toprule

        \multirow[c]{4}{*}{
            \makecell[c]{\footnotesize\textbf{Method}}
        }
        &
        \multirow[c]{3}{*}{
            \makecell[c]{
                \fontsize{7.5}{8.5}\selectfont\textbf{Encoder}\\
                \fontsize{7.5}{8.5}\selectfont\textbf{Params}\\
                \fontsize{7.5}{8.5}\selectfont\textbf{(M)}
            }
        }
        &
        \multicolumn{4}{!{\grayvrule}c!{\grayvrule}}{
            \footnotesize\bfseries Action Model
        }
        &
        \multicolumn{5}{c!{\grayvrule}}{
            \footnotesize\bfseries World Model
        }
        &
        \multicolumn{2}{c!{\grayvrule}}{
            \makecell[c]{
                \footnotesize\bfseries Inverse 
                \footnotesize\bfseries Dynamics
            }
        }
        &
        \multicolumn{3}{c}{
            \footnotesize\bfseries Reward Model
        }
        \\

        \cmidrule(lr){3-6}
        \cmidrule(lr){7-11}
        \cmidrule(lr){12-13}
        \cmidrule(l){14-16}

        &
        &
        \multicolumn{2}{!{\grayvrule}c!{\grayvrule}}{
            \scriptsize\bfseries Single-Frame
        }
        &
        \multicolumn{2}{c!{\grayvrule}}{
            \scriptsize\bfseries Sequence
        }
        &
        \multicolumn{2}{c!{\grayvrule}}{
            \scriptsize\bfseries Retrieval
        }
        &
        \multicolumn{3}{c!{\grayvrule}}{
            \scriptsize\bfseries Pixel-level
        }
        &
        \multirow[c]{2}{*}{
            \makecell[c]{\scriptsize T-MAE\\$\downarrow$}
        }
        &
        \multirow[c]{2}{*}{
            \makecell[c]{\scriptsize R-MAE\\$\downarrow$}
        }
        &
        \multirow[c]{2}{*}{
            \makecell[c]{\scriptsize Spearman\\$\rho\uparrow$}
        }
        &
        \multirow[c]{2}{*}{
            \makecell[c]{\scriptsize Kendall's\\$\tau\uparrow$}
        }
        &
        \multirow[c]{2}{*}{
            \makecell[c]{\scriptsize Rank\\Acc. $\uparrow$}
        }
        \\

        \cmidrule(lr){3-4}
        \cmidrule(lr){5-6}
        \cmidrule(lr){7-8}
        \cmidrule(lr){9-11}

        &
        &
        \makecell[c]{\scriptsize T-MAE\\$\downarrow$}
        &
        \makecell[c]{\scriptsize R-MAE\\$\downarrow$}
        &
        \makecell[c]{\scriptsize T-MAE\\$\downarrow$}
        &
        \makecell[c]{\scriptsize R-MAE\\$\downarrow$}
        &
        \makecell[c]{\scriptsize T-MAE\\$\downarrow$}
        &
        \makecell[c]{\scriptsize R-MAE\\$\downarrow$}
        &
        \makecell[c]{\scriptsize LPIPS\\$\downarrow$}
        &
        \makecell[c]{\scriptsize FID\\$\downarrow$}
        &
        \makecell[c]{\scriptsize SSIM\\$\uparrow$}
        & & & & &
        \\

        \midrule

        \multicolumn{16}{l}{
            \footnotesize\itshape
            Appearance-Centric Representations, Generic Pretrained Vision Foundation Models
        }
        \\

        \addlinespace[2pt]

        SigLIP 2~\cite{tschannen2025siglip}
        & 92.88
        & 18.58  &  21.14
        & 8.68 & 8.76
        & 18.14 & 22.83
        & 0.302 & 36.471 & 0.545
        & 25.71 & 32.94
        & 0.164 & 0.110 & 0.555
        \\

        DINOv3~\cite{simeoni2025dinov3}
        & 21.59
        & 16.90 & 19.67
        & 8.22 & 8.55
        & 12.90 & 16.51
        & 0.353 & 53.952 & 0.504
        & 23.44 & 30.89
        & 0.297 & 0.203 & 0.601
        \\

        \midrule

        \multicolumn{16}{l}{
            \footnotesize\itshape
            Appearance-Centric Representations, Pre-trained on Our Ultrasound Data
        }
        \\

        \addlinespace[2pt]

        US-MAE~\cite{mae}
        & 21.66
        & 17.48 & 19.34
        & 8.16 & 8.46
        & 14.99 & 18.52
        & 0.259 & \secondbest{31.180} & 0.586
        & 27.22 & 33.95
        & 0.206 & 0.140 & 0.568
        \\

        US-MoCo~\cite{mocov3}
        & 21.66
        & 16.79 & 19.39
        & 8.27 & 8.56
        & 15.30 & 20.19
        & 0.368 & 61.118 & 0.484
        & 25.60 & 32.25
        & 0.248 & 0.169 & 0.583
        \\

        \midrule

        \multicolumn{16}{l}{
            \footnotesize\itshape
            Appearance-Centric Representations, Open-source Medical and Ultrasound Foundation Models
        }
        \\

        \addlinespace[2pt]

        USFM~\cite{jiao2024usfm}
        & 85.67
        & 18.47 & 21.18
        & 8.23 & 8.51
        & 18.55 & 23.67
        & 0.312 & 46.777 & 0.532
        & 28.46 & 33.73
        & 0.136 & 0.092 & 0.545
        \\

        LVM-Med~\cite{mh2024lvm}
        & 85.74
        & 16.85 & 19.12
        & 8.12 & 8.55
        & 18.00 & 23.77
        & 0.369 & 71.125 & 0.483
        & 26.45 & 32.61
        & 0.345 & 0.238 & 0.618
        \\

        BioMedCLIP~\cite{zhang2023biomedclip}
        & 85.80
        & 16.13 & 19.30
        & 8.04 & 8.43
        & 13.83 & 18.97
        & 0.286 & 40.610 & 0.553
        & 22.90 & 30.59
        & 0.317 & 0.217 & 0.608
        \\

        EchoCLIP~\cite{christensen2024vision}
        & 87.57
        & 14.59 & 17.39
        & 7.75 & 8.20
        & 8.53 & 10.80
        & \secondbest{0.251}
        & 31.744
        & \secondbest{0.601}
        & 20.70 & 22.86
        & \secondbest{0.429}
        & \secondbest{0.300}
        & \secondbest{0.650}
        \\

        \midrule

        \multicolumn{16}{l}{
            \footnotesize\itshape
            Action-Conditioned Representations, Pre-trained on Our Ultrasound Data
        }
        \\

        \addlinespace[2pt]

        Structure-aware$^{\dag}$\cite{jiang2024structure}
        & 21.66
        & 16.78 & 18.42
        & 7.87 & 8.31
        & 13.62 & 16.79
        & 0.314 & 44.991 & 0.536
        & 23.75 & 27.44
        & 0.241 & 0.164 & 0.581
        \\

        EchoWorld$^{\dag}$~\cite{yue2025echoworld}
        & 21.66
        & 16.75 & 18.49
        & 7.92 & 8.29
        & 11.73 & 13.76
        & 0.287 & 44.168 & 0.565
        & 20.84 & 23.66
        & 0.233 & 0.159 & 0.578
        \\

        UltraSeP$^{\dag}$~\cite{jiang2025ultrasep}
        & 21.66
        & \secondbest{13.43}
        & \secondbest{14.21}
        & \secondbest{7.64}
        & \secondbest{7.96}
        & \secondbest{7.68}
        & \secondbest{8.21}
        & 0.274 & 36.536 & 0.582
        & \secondbest{19.38}
        & \secondbest{22.26}
        & 0.333 & 0.230 & 0.615
        \\

        \midrule

        \multicolumn{16}{l}{
            \footnotesize\itshape
            Action-Centric Representations, Pre-trained on Our Ultrasound Data
        }
        \\

        \addlinespace[2pt]

        \textbf{EchoACE (Ours)}
        & 21.66
        & \best{9.95}
        & \best{9.84}
        & \best{7.03}
        & \best{7.39}
        & \best{4.91}
        & \best{4.42}
        & \best{0.239}
        & \best{26.518}
        & \best{0.607}
        & \best{12.96}
        & \best{12.06}
        & \best{0.929}
        & \best{0.784}
        & \best{0.892}
        \\

        \multicolumn{2}{l!{\grayvrule}}{
            \gain{Gain over Best Baseline}
        }
        & \gain{$\downarrow$25.9\%}
        & \gain{$\downarrow$30.8\%}
        & \gain{$\downarrow$8.0\%}
        & \gain{$\downarrow$7.2\%}
        & \gain{$\downarrow$36.1\%}
        & \gain{$\downarrow$46.2\%}
        & \gain{$\downarrow$4.8\%}
        & \gain{$\downarrow$15.0\%}
        & \gain{$\uparrow$1.0\%}
        & \gain{$\downarrow$33.1\%}
        & \gain{$\downarrow$45.8\%}
        & \gain{$\uparrow$116.6\%}
        & \gain{$\uparrow$161.3\%}
        & \gain{$\uparrow$37.2\%}
        \\

        \bottomrule

    \end{tabularx}

    \endgroup

    \vspace{-5pt}

\end{table*}

% ============================================================
% Same-pose Output Consistency
% ============================================================

\begin{table}[t]
    \centering

    \caption{\textbf{Robustness to appearance variations at a fixed pose.}
    }
    \label{tab:same_pose_consistency}
    \vspace{-2mm}
    \resizebox{0.98\linewidth}{!}{%
    \small
    \begin{tabular}{@{}lcc}
        \toprule

        \textbf{Method}
        &
        \textbf{MAE of Trans. (mm)} $\downarrow$
        &
        \textbf{MAE of Rot. ($^\circ$)} $\downarrow$
        \\

        \midrule

        EchoCLIP
        & \secondbest{2.03 $\pm$ 0.02}
        & \secondbest{2.46 $\pm$ 0.02}
        \\

        UltraSeP
        & 2.53 $\pm$ 0.03
        & 3.13 $\pm$ 0.07
        \\

        \textbf{EchoACE (Ours)}
        & \best{1.06 $\pm$ 0.01 ($\downarrow$47.8\%)}
        & \best{1.40 $\pm$ 0.02 ($\downarrow$43.1\%)}
        \\

        \bottomrule

    \end{tabular}
    }

    \vspace{-3mm}

\end{table}

\begin{table}[t]
    \centering
    \caption{\textbf{Trajectory reconstruction errors.}
    }
    \vspace{-2mm}
    \label{tab:trajectory_reconstruction}
    \resizebox{0.98\linewidth}{!}{%
    \small
    \begin{tabular}{@{}lcc}
        \toprule
        \textbf{Method}
        &
        \textbf{Position RMSE (mm)} $\downarrow$
        &
        \textbf{Mean Rotation Error ($^\circ$)} $\downarrow$
        \\
        \midrule

        EchoCLIP
        & \secondbest{18.17}
        & 15.25
        \\

        UltraSeP
        & 19.07
        & \secondbest{14.07}
        \\

        \textbf{EchoACE (Ours)}
        & \best{16.88 ($\downarrow$7.1\%)}
        & \best{12.84 ($\downarrow$8.7\%)}
        \\

        \bottomrule
    \end{tabular}
    }
    % \vspace{-2mm}
    \vspace{-15pt}
\end{table}

\begin{table*}[t]

    \noindent
    \makebox[\textwidth][l]{%

    % ========================================================
    % LEFT: Fine-tuning Results + Pretraining Cost
    % ========================================================

    \begin{minipage}[t]{0.36\textwidth}
        \vspace{0pt}

        % ====================================================
        % TOP: Fine-tuning Results
        % ====================================================

        \begingroup

        \scriptsize

        \captionof{table}{
            \textbf{Fine-tuning results.}
            \ddag~indicates that only the last layers of the encoder are fine-tuned to equalize trainable parameters across all methods.
        }

        \label{tab:finetuning_results}

        \vspace{-2mm}

        % Compact table spacing
        \setlength{\tabcolsep}{1pt}
        \renewcommand{\arraystretch}{0.95}

        \begin{tabularx}{\linewidth}{
            @{}
            >{\raggedright\arraybackslash}p{0.38\linewidth}
            *{4}{>{\centering\arraybackslash}X}
            @{}
        }

            \toprule

            % ----------------------------------------
            % Header
            % ----------------------------------------

            \multirow{2}{*}{\textbf{Method}}
            &
            \multicolumn{2}{c}{
                \textbf{Action Model}
            }
            &
            \multicolumn{2}{c}{
                \shortstack[c]{
                    \textbf{Inverse Dynamics}
                }
            }
            \\

            \cmidrule(lr){2-3}
            \cmidrule(lr){4-5}

            &
            \shortstack[c]{T-MAE}
            &
            \shortstack[c]{R-MAE}
            &
            \shortstack[c]{T-MAE}
            &
            \shortstack[c]{R-MAE}
            \\

            \midrule

            % ----------------------------------------
            % Results
            % ----------------------------------------

            From Scratch
            & 10.34
            & 11.31
            & 13.02
            & 12.90
            \\

            EchoCLIP$^{\ddag}$
            & 8.23
            & \secondbest{8.69}
            & 12.56
            & 12.06
            \\

            UltraSeP
            & \secondbest{8.18}
            & 8.72
            & \secondbest{11.42}
            & \secondbest{11.18}
            \\

            \textbf{EchoACE (Ours)}
            & \best{7.83}
            & \best{8.13}
            & \best{10.80}
            & \best{10.35}
            \\

            \bottomrule

        \end{tabularx}

        \endgroup

        % Compact spacing between the two left tables
        \vspace{1mm}

        % ====================================================
        % BOTTOM: Pretraining Cost
        % ====================================================

        \captionof{table}{
            \textbf{Pretraining cost.}
                Structure-aware and EchoWorld use single-stage pretraining.
        }

        \label{tab:pretraining_cost}

        \vspace{1mm}

        \begingroup

        % Smaller font and compact row spacing
        \scriptsize

        \setlength{\tabcolsep}{2pt}
        \renewcommand{\arraystretch}{1.13}

        \begin{tabularx}{\linewidth}{
            @{}
            >{\raggedright\arraybackslash}p{0.34\linewidth}
            *{3}{>{\centering\arraybackslash}X}
            @{}
        }

            \toprule

            % ----------------------------------------
            % First header row
            % Method spans two rows
            % Pretraining Cost spans three columns
            % ----------------------------------------

            \multirow[c]{2}{*}{\textbf{Method}}
            &
            \multicolumn{3}{c}{
                \textbf{Pretraining Cost (GPU-hours)}
            }
            \\

            \cmidrule(lr){2-4}

            % ----------------------------------------
            % Second header row
            % ----------------------------------------

            &
            \textbf{Stage1}
            &
            \textbf{Stage2}
            &
            \textbf{Total}
            \\

            \midrule

            Structure-aware
            & /
            & /
            & \secondbest{61.0}
            \\

            EchoWorld
            & /
            & /
            & 120.0
            \\

            UltraSeP
            & \best{28.1}
            & \secondbest{171.0}
            & 199.1
            \\

            \textbf{EchoACE (Ours)}
            & \best{28.1}
            & \best{\textbf{0.6}}
            & \best{28.7}
            \\

            \bottomrule

        \end{tabularx}

        \endgroup

    \end{minipage}%

    \hfill%

    \begin{minipage}[t]{0.62\textwidth}
        \vspace{0pt}
    
        \captionof{table}{
            \textbf{Ablation studies.}
            We conduct two ablation studies: \textbf{(1) loss formulations}, comparing Direct Distance Regression, Learnable-Scale Regression, ListNet-style, and RankNet-style (Ours); and \textbf{(2) pretraining stages}, investigating the individual and joint contributions of Stage 1 and Stage 2 pretraining.
        }
    
        \label{tab:loss_ablation}
    
        \vspace{-2mm}
    
        \begingroup
    
        \scriptsize
        \setlength{\tabcolsep}{1pt}
        \renewcommand{\arraystretch}{1.2}
    
        \begin{tabularx}{\linewidth}{
            @{}
            >{\raggedright\arraybackslash}p{1.10cm}
            !{\color{gray!65}\vrule width 0.35pt}
            >{\raggedright\arraybackslash}p{1.85cm}
            *{8}{>{\centering\arraybackslash}X}
        }
    
            \toprule
    
            % ----------------------------------------
            % Header
            % ----------------------------------------
    
            \multicolumn{2}{c}{
                \multirow[c]{2}{*}{
                    \textbf{Ablation}
                }
            }
            &
            \multicolumn{2}{c}{
                \textbf{Action Model}
            }
            &
            \multicolumn{2}{c}{
                \textbf{World Model}
            }
            &
            \multicolumn{2}{c}{
                \textbf{Inv. Dynamics}
            }
            &
            \multicolumn{2}{c}{
                \textbf{Reward Model}
            }
            \\
    
            \cmidrule(lr){3-4}
            \cmidrule(lr){5-6}
            \cmidrule(lr){7-8}
            \cmidrule(lr){9-10}
    
            \multicolumn{2}{c}{}
            &
            \makecell[c]{\scriptsize T-MAE}
            &
            \makecell[c]{\scriptsize R-MAE}
            &
            \makecell[c]{\scriptsize T-MAE}
            &
            \makecell[c]{\scriptsize R-MAE}
            &
            \makecell[c]{\scriptsize T-MAE}
            &
            \makecell[c]{\scriptsize R-MAE}
            &
            \makecell[c]{\scriptsize Spear.}
            &
            \makecell[c]{\scriptsize Ken.}
            \\
    
            \midrule
            \midrule
    
            % ========================================
            % Section 1: Loss Formulations
            % ========================================
    
            \multicolumn{10}{@{}l}{
                \textbf{\textit{(1) Ablation study of different loss formulations in EchoACE}}
            }
            \\
    
            \midrule
    
            % ----------------------------------------
            % Regression
            % ----------------------------------------
    
            \multirow[c]{2}{*}{Regression}
            &
            \mbox{Direct Regression}
            & 17.61
            & 15.83
            & 19.03
            & 14.68
            & 27.27
            & 23.43
            & 0.546
            & 0.374
            \\
    
            &
            \mbox{Scaled Regression}
            & 16.68
            & \secondbest{14.97}
            & 10.30
            & 9.11
            & 24.90
            & 21.62
            & 0.593
            & 0.414
            \\
    
            \midrule
    
            % ----------------------------------------
            % Ranking
            % ----------------------------------------
    
            \multirow[c]{2}{*}{Ranking}
            &
            \mbox{ListNet}
            & \secondbest{15.25}
            & 15.51
            & \secondbest{5.90}
            & \secondbest{5.30}
            & \secondbest{20.37}
            & \secondbest{16.65}
            & \secondbest{0.791}
            & \secondbest{0.609}
            \\
    
            &
            \textbf{\mbox{RankNet (Ours)}}
            & \best{9.95}
            & \best{9.84}
            & \best{4.91}
            & \best{4.42}
            & \best{12.96}
            & \best{12.06}
            & \best{0.929}
            & \best{0.784}
            \\
    
            \midrule
            \midrule
    
            % ========================================
            % Section 2: Pretraining Stages
            % ========================================
    
            \multicolumn{10}{@{}l}{
                \textbf{\textit{(2) Ablation study of the contributions of Stage 1 and Stage 2 Pretraining in EchoACE}}
            }
            \\
    
            \midrule
    
            % ----------------------------------------
            % Stage 1 only
            % ----------------------------------------
    
            \multicolumn{2}{@{}l}{Stage1-only Pretrain}
            & 17.45
            & 19.65
            & 12.81
            & 15.89
            & 25.11
            & 27.08
            & 0.238
            & 0.162
            \\
    
            % ----------------------------------------
            % Stage 2 only
            % ----------------------------------------
    
            \multicolumn{2}{@{}l}{Stage2-only Pretrain}
            & \secondbest{13.34} & \secondbest{13.08} & \secondbest{9.38} & \secondbest{8.77} & \secondbest{20.13} & \secondbest{19.49} & \secondbest{0.754} & \secondbest{0.569}
            \\
    
            % ----------------------------------------
            % Stage 1 + Stage 2
            % ----------------------------------------
    
            \multicolumn{2}{@{}l}{\textbf{Stage1 + Stage2 Pretrain}}
            & \best{9.95}
            & \best{9.84}
            & \best{4.91}
            & \best{4.42}
            & \best{12.96}
            & \best{12.06}
            & \best{0.929}
            & \best{0.784}
            \\
    
            % ----------------------------------------
            % Best baseline for reference (gray)
            % ----------------------------------------
    
            \arrayrulecolor{gray!65}
            \specialrule{0.35pt}{2pt}{2pt}
            \arrayrulecolor{black}
    
            \multicolumn{2}{@{}l}{
                \textcolor{gray!65}{Best baseline (UltraSeP) for reference}
            }
            &
            \textcolor{gray!65}{13.43}
            &
            \textcolor{gray!65}{14.21}
            &
            \textcolor{gray!65}{7.68}
            &
            \textcolor{gray!65}{8.21}
            &
            \textcolor{gray!65}{19.38}
            &
            \textcolor{gray!65}{22.26}
            &
            \textcolor{gray!65}{0.333}
            &
            \textcolor{gray!65}{0.230}
            \\
    
            \bottomrule
    
        \end{tabularx}
    
        \endgroup
    
    \end{minipage}%

    }
    \vspace{-10pt}
\end{table*}

\subsection{Experimental Setup}

\noindent\textbf{Dataset.}
Our dataset comprises 356 expert scanning trajectories, split at the patient level into 256, 28, and 72 scans for training, validation, and testing, respectively, with test subjects held out from training. For Stage 2 pretraining, we divide physical distances into bins and uniformly sample within-scan pairs from each bin, obtaining 180,547 training and 19,453 validation pairs from the respective splits. For downstream task evaluation, the action model is evaluated on all 291,954 images from the 72 test scans. The world model, inverse dynamics, and reward model require pairwise evaluation, for which we sample 910 image pairs from each test scan by uniformly sampling within physical-distance bins, yielding 65,520 test pairs in total. See Supplementary~\ref{sec:supp_dataset} for further dataset details.

\noindent\textbf{Implementation Details.}
EchoACE Stage 1 follows I-JEPA pretraining for 50 epochs, while Stage 2 trains the encoder for 5 epochs with a learning rate of $3\times10^{-5}$, a batch size of 32, and a weight decay of 0.05. 
We compare EchoACE with four groups of baselines (Table~\ref{tab:result_main}), with baseline implementations detailed in Supplementary~\ref{sec:supp_baselines}. For evaluation across the four tasks, all methods use identical task-specific heads for fair comparisons. Evaluation details for each task are provided in Supplementary~\ref{sec:supp_downstream}.

\subsection{Performance on Four Tasks}
\noindent\textbf{Action Prediction for Probe Navigation.}
We report the mean absolute error (MAE) between predicted and ground-truth actions in translation (mm) and rotation ($^\circ$), averaged over ten standard views (Fig.~\ref{fig:dataset}). Per-view results are provided in the Supplementary~\ref{sec:supp_perview_action}.
We evaluate both single-frame and sequence inputs; within each setting, all methods share the same prediction head on top of their frozen encoders (details in the Supplementary~\ref{sec:supp_action}).
As shown in Table~\ref{tab:result_main}, EchoACE achieves the lowest errors, outperforming the strongest baseline by 25.9\%/30.8\% (trans./rot.) with single frames and 8.0\%/7.2\% with sequences.

\noindent\textit{Same-pose consistency.}
Navigation should be insensitive to appearance changes unrelated to probe motion, such as cardiac motion, respiration, and imaging parameters.
To evaluate this, we randomly sample 5,000 pairs of observations at the same probe pose and compute the MAE between the predicted actions of the two images in each pair, averaging across pairs. This procedure is repeated three times in the single-frame setting.
As shown in Table~\ref{tab:same_pose_consistency}, EchoACE reduces this discrepancy to 1.06\,mm and 1.40$^\circ$, which is 47.8\% and 43.1\% lower than the best baseline.
These results show that EchoACE is more robust to non-physical appearance changes, yielding more stable guidance signals.

\begin{figure}[!t]
    \centering
    \includegraphics[width=\linewidth]{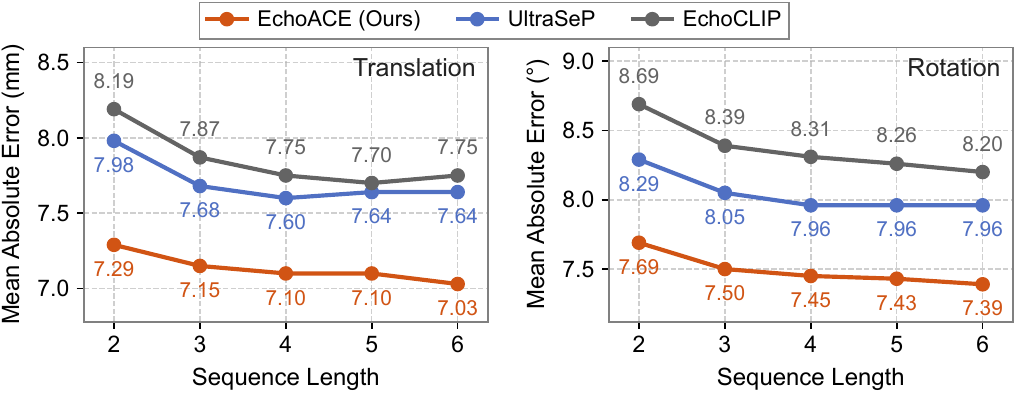}
    \caption{\textbf{Action Prediction Error vs. Input Sequence Length.}}
    \label{fig:seq_len}
    \vspace{-5pt}
\end{figure}

\noindent\textit{Effect of sequence length.}
As shown in Fig.~\ref{fig:seq_len}, EchoACE consistently outperforms the baselines across all sequence lengths from 2 to 6.
Notably, with only two frames, it already surpasses both baselines with six frames.

\noindent\textbf{Action-Conditioned World Modeling.}
We evaluate the predicted future representations at two levels.
At the latent level, each predicted latent is used to retrieve its nearest neighbor from the dataset, and we report the pose MAE between the retrieved and ground-truth target observations.
At the pixel level, we train a diffusion decoder for each method to map predicted latents to images (details in the Supplementary~\ref{sec:supp_world_model}), and report LPIPS, FID, and SSIM against the ground truth.
As shown in Table~\ref{tab:result_main}, EchoACE reduces the retrieval errors by 36.1\%/46.2\% (translation/rotation) relative to the strongest baseline, indicating that its predicted latents more accurately reflect the physical outcomes of probe actions.
It also achieves the best results on all pixel-level metrics.
Qualitatively, EchoACE produces predictions closer to the ground truth in both one-step prediction and autoregressive rollout (Fig.~\ref{fig:world_model}), suggesting its potential as a learned simulator for training novice sonographers.

\noindent\textbf{Inverse Dynamics Prediction.}
We report the MAE between the predicted and ground-truth relative actions of two observations.
As shown in Table~\ref{tab:result_main}, EchoACE reduces the errors by 33.1\%/45.8\% (translation/rotation) relative to the strongest baseline.
We further reconstruct probe trajectories by chaining the relative poses predicted between consecutive frames, using~\cite{Lategahn2012IV} to mitigate error accumulation (details in the Supplementary~\ref{sec:supp_trajectory}).
On 4,000 trajectories of 100 frames each, EchoACE reduces the position RMSE and mean rotation error by 7.1\% and 8.7\% over the best baseline (Table~\ref{tab:trajectory_reconstruction}), and its reconstructed trajectories more closely follow the ground truth (Fig.~\ref{fig:recon}(a)).

\noindent\textit{Left ventricle (LV) 3D reconstruction.}
Using the poses estimated by the inverse dynamics model, we place LV masks from ten views into a common 3D space to reconstruct the LV (Fig.~\ref{fig:recon}(b); details in Supplementary~\ref{sec:supp_3d}).
This demonstrates the potential of EchoACE for sensorless 3D cardiac reconstruction, which is valuable for assessing LV function.

\begin{figure}[!t]
    \centering
    \includegraphics[width=\linewidth]{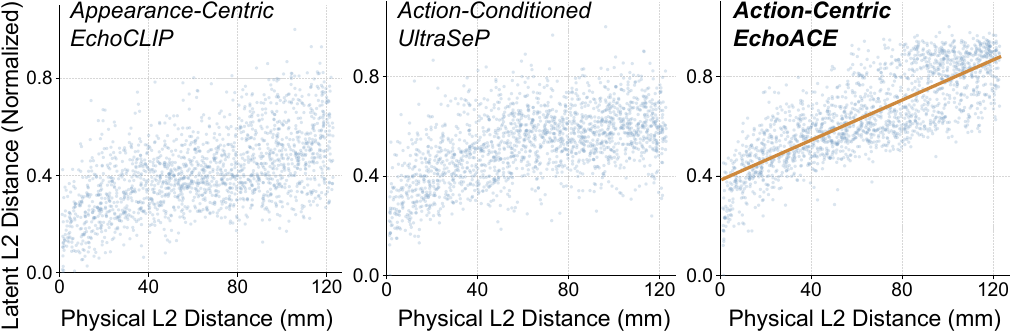}
    \caption{\textbf{Latent vs. physical distance} (1794 samples).}
    \label{fig:sandian}
    \vspace{-15pt}
\end{figure}

\begin{figure*}[!t]
    \centering
    \includegraphics[width=\linewidth]{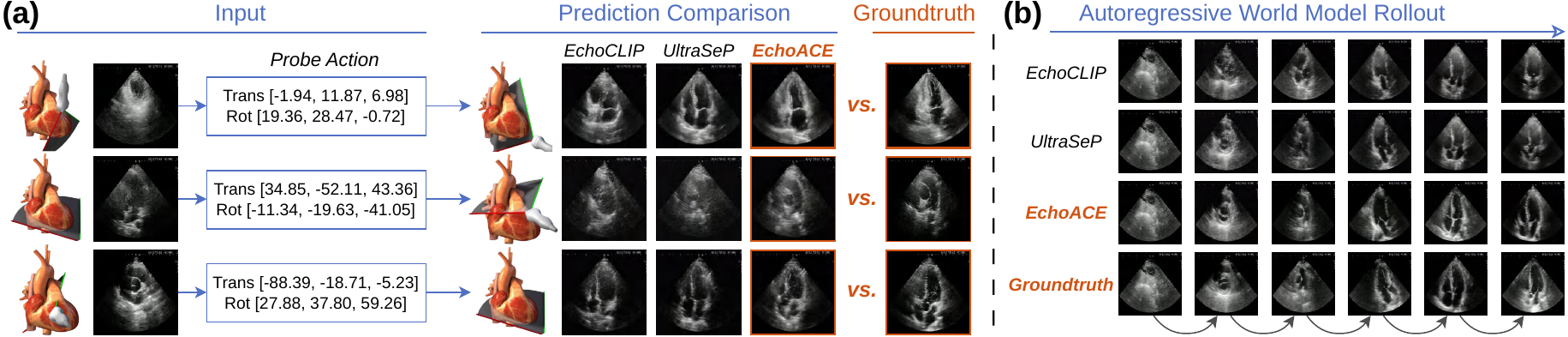}
    \caption{\textbf{World model visualization}.
    (a) One-step world model prediction. 
    (b) Autoregressive world model rollout.
    }
    \label{fig:world_model}
    \vspace{-5pt}
\end{figure*}

\begin{figure*}[!t]
    \centering
    \includegraphics[width=\linewidth]{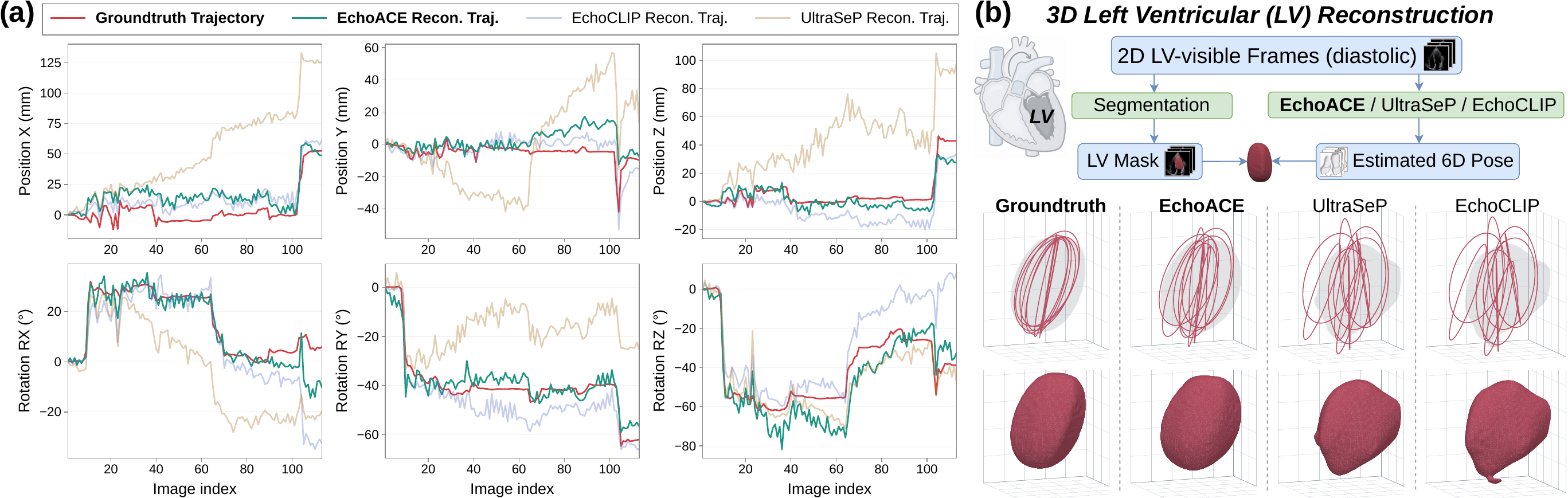}
    \caption{\textbf{Reconstruction results}.
    (a) Trajectory reconstruction. 
    (b) 3D reconstruction of the left ventricle.
    }
    \label{fig:recon}
    \vspace{-5pt}
\end{figure*}

\noindent\textbf{Latent Distance as a Progress Reward.}
We examine whether the latent distance to a goal observation can directly serve as a progress reward without training a reward network.
We measure the agreement between latent and physical distances using Spearman's $\rho$, Kendall's $\tau$, and pairwise ranking accuracy.
As shown in Table~\ref{tab:result_main}, compared with the strongest baseline, EchoACE raises $\rho$ from 0.43 to 0.93, $\tau$ from 0.30 to 0.78, and ranking accuracy from 0.65 to 0.89.
Consistently, its latent distances increase monotonically with physical distances (Fig.~\ref{fig:sandian}), whereas those of the baselines are only weakly correlated.

\begin{figure*}[!t]
    \centering
    \includegraphics[width=\linewidth]{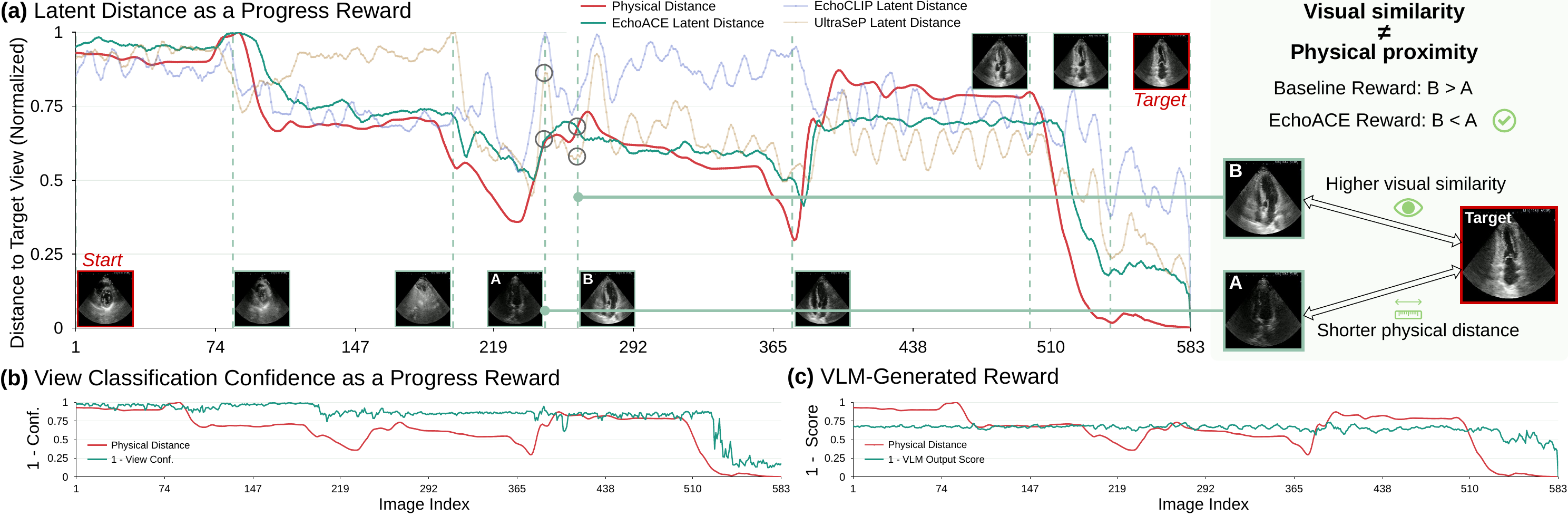}
    \caption{\textbf{Progress estimation in a real scanning sequence.} 
    (a) Latent distance as a progress reward. 
    (b) View classification confidence as a progress reward.
    (c) VLM-generated reward. 
    }
    \label{fig:vis_reward}
    \vspace{-5pt}
\end{figure*}

On a real scanning sequence (Fig.~\ref{fig:vis_reward}(a)), the latent distance of EchoACE closely tracks the physical distance to the target view throughout the trajectory.
In contrast, baseline latent distances fluctuate and can be misled by visual similarity, assigning a higher reward to a visually similar but physically distant view (B) than to a physically closer one (A).
Moreover, view classification confidence and VLM-generated rewards (Fig.~\ref{fig:vis_reward}(b,c); details in the Supplementary~\ref{sec:supp_reward_baselines}) remain nearly constant until the target is reached, offering little guidance.

\noindent\textbf{Fine-tuning Results.}
Beyond the frozen-encoder setting, we further fine-tune the encoders on the action prediction and inverse dynamics tasks (Table~\ref{tab:finetuning_results}).
EchoACE still achieves the lowest errors, outperforming the strongest baseline by 4.3--7.4\% and training from scratch by up to 28.1\%, indicating that it also provides a better initialization for downstream adaptation.

\subsection{Ablation Studies}
\noindent\textbf{Loss formulations.}
As shown in Table~\ref{tab:loss_ablation}~(1), ranking objectives consistently outperform regression objectives.
Both direct and scaled regression yield substantially higher errors.
Among ranking objectives, our RankNet-style formulation further outperforms the ListNet-style formulation~\cite{cao2007learning}.
We attribute the inferiority of regression to inter-subject anatomical variability: since absolute distances between views vary across patients, they form an inconsistent regression target, whereas their relative ordering remains consistent and thus provides more reliable supervision.

\noindent\textbf{Pretraining stages.}
As shown in Table~\ref{tab:loss_ablation}~(2), Stage 1 alone yields an appearance-centric encoder with weak physical awareness (Spearman $\rho$ of 0.24).
Stage 2 alone substantially improves physical awareness ($\rho$ of 0.75), comparable to the strongest baseline on downstream tasks. Combining both stages further achieves the best results across all tasks, indicating that the visual prior and physical distance supervision are complementary.

\noindent\textbf{Pretraining cost.}
As shown in Table~\ref{tab:pretraining_cost}, EchoACE's action-centric pretraining (Stage 2) takes only 0.6 GPU-hours versus 171.0, a 285$\times$ reduction.
In terms of total cost, EchoACE requires only 28.7 GPU-hours versus 199.1 for UltraSeP, the strongest baseline, a 6.9$\times$ reduction.
All GPU-hour measurements are based on NVIDIA H20 GPUs.

\section{Conclusion}
In this paper, we presented EchoACE, an action-centric foundation model for embodied echocardiography.
Observing that existing appearance-centric representations fail to preserve the physical proximity of ultrasound views, EchoACE introduces a Physical Distance Ranking objective that directly aligns the ordering of latent distances with that of physical probe displacements.
The resulting shared representation supports four complementary embodied capabilities, namely probe navigation, world modeling, inverse dynamics prediction, and progress estimation, consistently outperforming both appearance-centric and action-conditioned baselines, including larger echocardiography foundation models, at a substantially lower pretraining cost.
While our evaluation follows the offline protocols of prior work, future work will integrate EchoACE into robotic echocardiography systems for closed-loop validation.
\section{Acknowledgments}
This work is supported in part by the National Key R\&D Program of China under Grant 2024YFB4708200 and the National Natural Science Foundation of China under Grants U24B20173, U2541227 and 62625604.

{
    \small
    \bibliographystyle{ieeenat_fullname}
    \bibliography{main}
}

\clearpage
\maketitlesupplementary

\setcounter{section}{0}
\setcounter{subsection}{0}

% Reset figure, table, and equation counters
\setcounter{figure}{0}
\setcounter{table}{0}
\setcounter{equation}{0}

% Supplementary section numbering
\renewcommand{\thesection}{\Alph{section}}
\renewcommand{\thesubsection}{\thesection.\arabic{subsection}}
\renewcommand{\thesubsubsection}{\thesubsection.\arabic{subsubsection}}

% Supplementary figure, table, and equation numbering
\renewcommand{\thefigure}{S\arabic{figure}}
\renewcommand{\thetable}{S\arabic{table}}
\renewcommand{\theequation}{S\arabic{equation}}

% Hyperref destination numbering
\renewcommand{\theHsection}{supp.\Alph{section}}
\renewcommand{\theHsubsection}{\theHsection.\arabic{subsection}}
\renewcommand{\theHsubsubsection}{\theHsubsection.\arabic{subsubsection}}

% ============================================================
% Table of Contents (Supplementary only)
% ============================================================

\begin{center}
    \Large\bfseries Contents
\end{center}

\vspace{3mm}

\noindent
\hyperref[sec:supp_implementation]{\textbf{A. Implementation Details}}
\dotfill
\pageref{sec:supp_implementation}

\hspace{1em}
\hyperref[sec:supp_dataset]{A.1 Dataset}
\dotfill
\pageref{sec:supp_dataset}

\hspace{1em}
\hyperref[sec:supp_echoace_pretraining]{A.2 EchoACE Pretraining Details}
\dotfill
\pageref{sec:supp_echoace_pretraining}

\hspace{1em}
\hyperref[sec:supp_baselines]{A.3 Baseline Implementations}
\dotfill
\pageref{sec:supp_baselines}

\vspace{2mm}

\noindent
\hyperref[sec:supp_downstream]{\textbf{B. Downstream Task Protocols}}
\dotfill
\pageref{sec:supp_downstream}

\hspace{1em}
\hyperref[sec:supp_action]{B.1 Action Prediction for Probe Navigation}
\dotfill
\pageref{sec:supp_action}

\hspace{1em}
\hyperref[sec:supp_world_model]{B.2 Action-Conditioned World Modeling}
\dotfill
\pageref{sec:supp_world_model}

\hspace{1em}
\hyperref[sec:supp_inverse]{B.3 Inverse Dynamics Prediction}
\dotfill
\pageref{sec:supp_inverse}

\hspace{1em}
\hyperref[sec:supp_reward]{B.4 Latent Distance as a Progress Reward}
\dotfill
\pageref{sec:supp_reward}

\vspace{2mm}

\noindent
\hyperref[sec:supp_additional_eval]{\textbf{C. Additional Evaluation Protocols}}
\dotfill
\pageref{sec:supp_additional_eval}

\hspace{1em}
\hyperref[sec:supp_trajectory]{C.1 Probe Trajectory Reconstruction}
\dotfill
\pageref{sec:supp_trajectory}

\hspace{1em}
\hyperref[sec:supp_3d]{C.2 Left-Ventricular 3D Reconstruction}
\dotfill
\pageref{sec:supp_3d}

\hspace{1em}
\hyperref[sec:supp_reward_baselines]{C.3 Progress Reward Baselines in Figure 9}
\dotfill
\pageref{sec:supp_reward_baselines}

\vspace{2mm}

\noindent
\hyperref[sec:supp_results]{\textbf{D. Additional Results}}
\dotfill
\pageref{sec:supp_results}

\hspace{1em}
\hyperref[sec:supp_reference_length]{D.1 Sensitivity to the Reference Length \(L\)}
\dotfill
\pageref{sec:supp_reference_length}

\hspace{1em}
\hyperref[sec:supp_perview_action]{D.2 Per-View Action Prediction Errors}
\dotfill
\pageref{sec:supp_perview_action}

\hspace{1em}
\hyperref[sec:supp_external_val]{D.3 Cross-Device External Validation}
\dotfill
\pageref{sec:supp_external_val}

\section{Implementation Details}
\label{sec:supp_implementation}

\subsection{Dataset}
\label{sec:supp_dataset}

\paragraph{Data Acquisition.}
Two senior sonographers with over 10 years of experience performed continuous scans covering ten standard cardiac views (Fig.~\ref{fig:dataset}) using an ultrasound probe mounted on a robotic arm. Data were acquired using a GE Vivid E7 ultrasound device (General Electric, USA) equipped with an M5S probe at a frame rate of 30 Hz. 
During scanning, the sonographers adjusted imaging parameters as in routine clinical practice, introducing natural variations in image appearance. Throughout the acquisition, ultrasound images and corresponding six-dimensional probe poses were synchronously recorded, and the ten standard cardiac views were annotated by the sonographers.
Each observation is represented as $(\mathbf{x}_t,\mathbf{T}_t)$, where $\mathbf{x}_t\in\mathbb{R}^{3\times H\times W}$ denotes the ultrasound image and $\mathbf{T}_t\in SE(3)$ denotes its corresponding probe pose.
The data collection process was conducted in compliance with ethical guidelines and was reviewed and approved by the relevant institutional ethics committee.

\paragraph{Dataset Splits.}
We collected data from 178 subjects, with two expert scanning trajectories acquired per subject, yielding 356 trajectories and 1,316,464 synchronized image--pose pairs. We split the dataset at the patient level into training, validation, and test sets, ensuring that no subject appears across different splits. The resulting splits contain 128, 14, and 36 subjects, respectively, corresponding to 256, 28, and 72 scans.

\begin{table}[t]
    \centering
    \caption{\textbf{Dataset statistics.}}
    \label{tab:supp_dataset_split}
    \vspace{-2mm}
    \begin{tabular}{lcccc}
        \toprule
        Split & Subjects & Scans & Images & Pairs \\
        \midrule
        Training & 128 & 256 & 922,858 & 180,547 \\
        Validation & 14 & 28 & 101,652 & 19,453 \\
        Test & 36 & 72 & 291,954 & 65,520 \\
        \midrule
        Total & 178 & 356 & 1,316,464 & 265,520 \\
        \bottomrule
    \end{tabular}
    \vspace{-5pt}
\end{table}

\paragraph{Pair Construction and Sampling.}

EchoACE Stage~2 pretraining and the downstream world model, inverse dynamics, and reward model require image pairs with corresponding relative probe motions. We therefore construct within-scan pairs, ensuring that both observations in each pair belong to the same scanning trajectory.

\textbf{Relative Motion Computation and Binning.}
Each frame is associated with probe-pose labels relative to the standard cardiac views available in its scan. For a candidate pair $(i,j)$, we select a standard view $k$ for which both frames have valid pose labels. Let $\mathbf{T}_i^{(k)}$ and $\mathbf{T}_j^{(k)}$ denote their corresponding homogeneous probe transformations with respect to the selected common reference view. Following the convention in the main paper, the relative transformation is computed as:
\begin{equation*}
\Delta\mathbf{T}_{i\rightarrow j}
=
\left(\mathbf{T}_i^{(k)}\right)^{-1}\mathbf{T}_j^{(k)}
=
\begin{bmatrix}
\Delta\mathbf{R}_{ij} & \Delta\mathbf{t}_{ij}\\
\mathbf{0}^{\top} & 1
\end{bmatrix}.
\end{equation*}
Here, $\Delta\mathbf{R}_{ij}\in SO(3)$ and $\Delta\mathbf{t}_{ij}\in\mathbb{R}^3$ denote the relative rotation and translation, respectively. We represent the rotation using its axis-angle vector $\boldsymbol{\phi}_{ij}=(\operatorname{Log}(\Delta\mathbf{R}_{ij}))^\vee$. We characterize the relative motion using its translation magnitude and rotation angle:
\begin{equation*}
d_t(i,j)=\|\Delta\mathbf{t}_{ij}\|_2,\quad
d_r(i,j)=\frac{180}{\pi}\|\boldsymbol{\phi}_{ij}\|_2.
\end{equation*}
We divide translation (mm) into seven bins: $[0,1)$, $[1,2)$, $[2,5)$, $[5,10)$, $[10,20)$, $[20,40)$, and $[40,+\infty)$, and rotation (degrees) into seven bins: $[0,2)$, $[2,5)$, $[5,10)$, $[10,20)$, $[20,40)$, $[40,80)$, and $[80,181)$. Their Cartesian product forms a $7\times7$ grid of 49 motion cells.

\textbf{Offline Pair Pool Construction.}
Training and validation scans are separated before pair generation, and their pair pools are constructed independently. To generate candidate pairs, we first sample a scan uniformly and then select an anchor frame uniformly within that scan. The second frame is sampled using a mixture of temporal sampling strategies: with probability 0.55, its temporal offset follows an approximately log-uniform distribution, favoring local motions; with probability 0.45, it is sampled uniformly from the remaining frames to provide broader temporal coverage. 

Each valid pair is assigned to one of the 49 motion cells. To control the size of the candidate pool, we maintain an independent fixed-capacity reservoir for each cell using standard reservoir sampling. We then select unique pairs from non-empty cells in a round-robin manner, subject to a per-scan sampling cap. If a cell contains fewer candidates than its desired allocation, all available pairs are retained without duplication. This procedure provides coverage across different motion ranges while limiting the contribution of individual scans. It yields 180,547 training pairs and 19,453 validation pairs.

\textbf{Training-Time Sampling.}
During Stage~2 pretraining, we further apply a capped square-root reweighting strategy to balance the contribution of different motion cells. Let $n_c$ denote the number of unique training pairs in a cell $c$, and $n_{\mathrm{med}}$ the median pair count across all cells. We define the cell reweighting factor as:
\begin{equation*}
m_c=
\operatorname{clip}
\left(
\sqrt{\frac{n_{\mathrm{med}}}{n_c}},
0.5,3.0
\right).
\end{equation*}
The sampling mass of each cell is $M_c=n_cm_c$. Given $N_{\mathrm{epoch}}$ samples per epoch, its ideal sampling quota is:
\begin{equation*}
q_c^*
=
N_{\mathrm{epoch}}
\frac{M_c}{\sum_k M_k}.
\end{equation*}
The ideal quotas $q_c^*$ are converted into integer quotas $q_c$ using largest-remainder allocation and capped at $3n_c$, limiting the use of each unique pair to at most three times per epoch.

At the beginning of each epoch, pair indices within each cell are randomly shuffled and stored in independent queues. Pairs are sampled without replacement within each pass, with the queue reshuffled only after exhaustion. The sampled pairs from all cells are then combined, globally shuffled, and distributed across training workers. The sampling order is reproducible but changes across epochs through epoch-dependent random seeds.

\textbf{Downstream Test Set.}
For downstream evaluation, we construct a fixed test set by uniformly sampling 910 within-scan pairs from each of the 72 held-out test scans across the motion cells, yielding 65,520 test pairs in total. No pair is duplicated, and the same test set is used for all compared methods without dynamic resampling.

\subsection{EchoACE Pretraining Details}
\label{sec:supp_echoace_pretraining}

EchoACE adopts a two-stage pretraining pipeline. Stage 1 establishes a general visual prior of cardiac ultrasound, while Stage 2 incorporates physical probe displacement supervision to explicitly structure the latent geometry.

\paragraph{Stage 1: Visual Prior Pretraining.}

We follow the I-JEPA-style paradigm~\cite{ijepa} to pretrain the image encoder on the training split. Specifically, visible context patches are encoded by a context encoder, while an exponential moving average (EMA) target encoder extracts representations from the full image. A lightweight predictor takes the context features and positional mask tokens to predict the representations of randomly sampled target blocks. The encoder is optimized through feature prediction rather than pixel-level image reconstruction. The main pretraining configurations are summarized in Table~\ref{tab:supp_stage1_config}.

\begin{table}[t]
    \centering
    \caption{\textbf{EchoACE Stage 1 pretraining configuration.}}
    \label{tab:supp_stage1_config}
    \vspace{-2mm}
    \small
    \begin{tabular}{lc}
        \toprule
        Configuration & Value \\
        \midrule
        \multicolumn{2}{c}{\textit{Architecture}} \\
        \midrule
        Encoder & ViT-Small \\
        Patch size & 16 \\
        Input resolution & $224\times224$ \\
        Predictor depth & 6 \\
        Predictor embedding dimension & 384 \\
        Number of context masks & 1 \\
        Number of target masks & 4 \\
        Context mask scale & $[0.85,1.0]$ \\
        Target mask scale & $[0.15,0.20]$ \\
        Target aspect ratio & $[0.75,1.5]$ \\
        \midrule
        \multicolumn{2}{c}{\textit{Training}} \\
        \midrule
        Training epochs & 50 \\
        Batch size & 256 \\
        Optimizer & AdamW \\
        Learning rate schedule & Cosine decay \\
        Initial learning rate & $2\times10^{-4}$ \\
        Weight decay & $0.04\rightarrow0.4$ \\
        EMA decay & $0.996\rightarrow1.0$ \\
        Numerical precision & FP32 \\
        \bottomrule
    \end{tabular}
\end{table}

\begin{table}[t]
    \centering
    \caption{\textbf{EchoACE Stage 2 pretraining configuration.}}
    \label{tab:supp_stage2_config}
    \vspace{-2mm}
    \small
    \begin{tabular}{lc}
        \toprule
        Configuration & Value \\
        \midrule
        Training epochs & 5 \\
        Batch size & 32 \\
        Optimizer & AdamW \\
        Learning rate & $3\times10^{-5}$ \\
        Weight decay & 0.05 \\
        Learning rate schedule & cosine decay \\
        Ranking temperature ($\tau$) & 0.1 \\
        Reference length ($L$) & 100 mm \\
        \bottomrule
    \end{tabular}
    \vspace{-5pt}
\end{table}

\paragraph{Stage 2: Action-Centric Pretraining.}

Starting from the Stage 1 initialization, we further optimize the encoder using the Physical Distance Ranking objective. We use 180,547 training pairs to optimize the encoder and 19,453 validation pairs for checkpoint selection. Details of the pair data are provided in Section~\ref{sec:supp_dataset}. The Stage 2 pretraining configuration is summarized in Table~\ref{tab:supp_stage2_config}.

\subsection{Baseline Implementations}
\label{sec:supp_baselines}

We compare EchoACE with four groups of baselines, as summarized in Table~\ref{tab:result_main}: (1) generic vision foundation models, including SigLIP 2 and DINOv3; (2) classical self-supervised methods, including MAE and MoCo; (3) open-source medical and ultrasound foundation models, including USFM, LVM-Med, BioMedCLIP, and EchoCLIP; and (4) action-conditioned pretraining methods, including Structure-aware, EchoWorld, and UltraSeP. All methods use an input resolution of $224\times224$ and follow the same downstream evaluation protocols, with identical task-specific prediction heads within each task.

\paragraph{Generic Vision Foundation Models.}

SigLIP 2~\cite{tschannen2025siglip} is a vision-language foundation model that learns visual representations through image-text pretraining using a sigmoid-based contrastive objective. We use its official \texttt{siglip2-base-patch16-224} checkpoint and extract features using its visual encoder. DINOv3~\cite{simeoni2025dinov3} is a self-supervised vision foundation model pretrained on large-scale general-domain images. We use its official \texttt{ViT-S/16 distilled} checkpoint.

\paragraph{Classical Self-Supervised Methods.}

We implement MAE~\cite{mae} and MoCo~\cite{mocov3} on our ultrasound training data, using ViT-Small encoders with 21.66M parameters. MAE learns visual representations by reconstructing randomly masked image patches. We follow its standard pretraining paradigm with a masking ratio of 0.75 and train for 300 epochs using AdamW, a learning rate of $6\times10^{-4}$, a weight decay of 0.05, and a batch size of 1024. MoCo learns representations through contrastive learning with a momentum encoder. We adopt the MoCo v3 framework and train for 150 epochs using AdamW, a learning rate of $1.5\times10^{-4}$, a weight decay of 0.1, and a batch size of 1024.

\paragraph{Open-Source Medical Foundation Models.}

We additionally evaluate four publicly available medical and ultrasound foundation models using their official pretrained weights. BioMedCLIP~\cite{zhang2023biomedclip} is a multimodal biomedical foundation model pretrained on 15 million biomedical image-text pairs through contrastive learning. LVM-Med~\cite{mh2024lvm} employs a graph-matching formulation for self-supervised representation learning, integrating diverse medical imaging modalities, including ultrasound. USFM~\cite{jiao2024usfm} is an ultrasound-specific foundation model pretrained on over two million ultrasound images using spatial-frequency dual masked image modeling. EchoCLIP~\cite{christensen2024vision} is a multimodal foundation model for echocardiogram interpretation, pretrained on over one million echocardiographic image-text pairs using contrastive learning. We use the visual encoder of each model to extract image representations.

\paragraph{Action-Conditioned Pretraining Methods.}

We compare EchoACE with three action-conditioned methods: Structure-aware~\cite{jiang2024structure}, EchoWorld~\cite{yue2025echoworld}, and UltraSeP~\cite{jiang2025ultrasep}. Structure-aware and EchoWorld use probe actions as conditioning inputs for visual prediction, while UltraSeP adopts masked vision-action sequence prediction, jointly predicting masked image features and probe movement actions. Despite their different predictive objectives, all three methods rely on additional prediction modules to learn from action-related supervision, rather than directly structuring the encoder's latent geometry according to physical probe displacements.
We follow the original pretraining configurations of these methods and train them on our ultrasound training data. Structure-aware is pretrained for 50 epochs and EchoWorld for 300 epochs. UltraSeP adopts a two-stage pipeline: its first stage is initialized with the same I-JEPA-pretrained weights used by EchoACE, followed by sequence-aware pretraining for 50 epochs in the second stage. For all three methods, we retain the pretrained visual encoder for downstream evaluation.

\section{Downstream Task Protocols}
\label{sec:supp_downstream}

\subsection{Action Prediction for Probe Navigation}
\label{sec:supp_action}

We evaluate Action Model under two settings: single-frame input and sequence input. Both settings use the pretrained encoder followed by a task-specific action prediction head.

\paragraph{Single-Frame Architecture.}

Given an ultrasound image $\mathbf{x}_t$, the pretrained encoder extracts its global representation:
\begin{equation*}
\mathbf{z}_t=f_\theta(\mathbf{x}_t).
\end{equation*}
We attach ten independent MLP heads to the encoder, one for each standard view. Each head consists of two fully connected layers with a GELU activation between them and predicts a 6D relative probe movement toward its corresponding target view. The prediction for target view $k$ is:
\begin{equation*}
\hat{\mathbf{a}}_{t\rightarrow k}
=
h_{\phi_k}(\mathbf{z}_t),
\quad k\in\{1,\ldots,10\}.
\end{equation*}
The prediction heads are trained using Smooth L1 loss between the predicted and ground-truth actions.

\paragraph{Sequence Architecture.}

\begin{figure}[t]
    \centering
    \includegraphics[width=\linewidth]{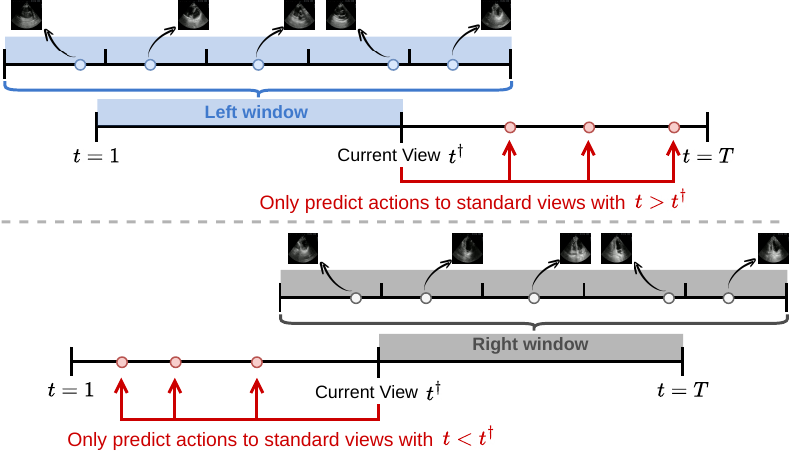}
    \caption{
        \textbf{Bidirectional segmental sampling for the sequence-input Action Model.}
        Given a current observation, input images are sampled from either the left or right temporal window, while action prediction targets are selected from the opposite side.
    }
    \label{fig:supp_sample}
    \vspace{-13pt}
\end{figure}

To evaluate the model's ability to leverage broader scanning context for action prediction, we additionally consider a sequence-based setting. Following the bidirectional sampling strategy of UltraSeP~\cite{jiang2025ultrasep} and EchoWorld~\cite{yue2025echoworld}, we construct input sequences from both temporal directions around the current observation, as illustrated in Fig.~\ref{fig:supp_sample}.

This design reflects the non-fixed order of clinical echocardiographic scanning, where physicians may revisit previously passed views after acquiring information from other cardiac planes. By considering observations from both sides of the current view, the model can exploit broader scanning context to support navigation toward remaining standard views.
Specifically, given a current observation at timestep $t^\dagger$ and an input sequence length $T$, we consider two complementary sampling directions. For left-window sampling, we divide the preceding scanning window into $T-1$ temporal segments and randomly sample one image from each segment. Together with the current observation, these images form an input sequence of length $T$. The model is supervised to predict relative actions toward standard views occurring on the right side of the current observation ($t>t^\dagger$). Conversely, for right-window sampling, we sample $T-1$ images from the following scanning window using the same segmental strategy and predict relative actions toward standard views on the left side ($t<t^\dagger$). This bidirectional segmental sampling promotes broader temporal coverage and captures more diverse cardiac structures along the scanning trajectory.

The relative probe actions between consecutive images in the sampled sequence are computed from their pose annotations. Each image is independently encoded by the pretrained encoder, while each relative action is embedded through a linear projection. Time-step positional embeddings are added to both image and action embeddings to encode their temporal positions. The resulting embeddings are processed by a four-layer Transformer to model temporal dependencies, using a hidden dimension of 192, four attention heads, and an MLP ratio of 2. Finally, ten independent two-layer MLP heads predict the relative movements toward the ten standard views. The action loss is computed only for target views on the opposite temporal side of the current observation. The Transformer and MLP heads jointly form the sequence-based action prediction head, which is trained using Smooth L1 loss between predicted and ground-truth actions.

\paragraph{Training Details.}

Both action prediction settings are trained for five epochs using AdamW, a batch size of 256, a learning rate of $10^{-4}$ and cosine learning rate scheduling. By default, the pretrained encoder is frozen, and only the task-specific head is optimized. The training configurations are summarized in Table~\ref{tab:supp_action_config}.

\begin{table}[t]
    \centering
    \caption{\textbf{Action Model training configuration.}}
    \label{tab:supp_action_config}
    \vspace{-2mm}
    \small
    \begin{tabular}{lc}
        \toprule
        Configuration & Value \\
        \midrule
        \multicolumn{2}{c}{\textit{Single-Frame Architecture}} \\
        \midrule
        Number of MLP heads & 10 \\
        MLP layers & 2 \\
        Activation & GELU \\
        \midrule
        \multicolumn{2}{c}{\textit{Sequence Architecture}} \\
        \midrule
        Default sequence length & 6 \\
        Transformer depth & 4 blocks \\
        Hidden dimension & 192 \\
        Attention heads & 4 \\
        MLP ratio & 2 \\
        Number of MLP heads & 10 \\
        MLP layers & 2 \\
        Activation & GELU \\
        Action embedding & Linear \\
        \midrule
        \multicolumn{2}{c}{\textit{Training}} \\
        \midrule
        Training epochs & 5 \\
        Batch size & 256 \\
        Optimizer & AdamW \\
        Loss & Smooth L1 \\
        Initial learning rate & $10^{-4}$ \\
        Learning rate schedule & Cosine \\
        \bottomrule
    \end{tabular}
    \vspace{-10pt}
\end{table}

\paragraph{Evaluation Protocol and Metrics.}

We evaluate the action model on all ultrasound images from the 72 held-out test scans. For each input image, the model predicts 6D relative actions toward all ten standard views. The first three dimensions represent translation (mm), while the last three represent Euler angles ($^\circ$).
We report translation and rotation mean absolute errors (T-MAE and R-MAE), averaged over all test observations and ten target views. Let $N_a$ denote the number of test observations and $K=10$ the number of target views. The metrics are defined as:
\begin{equation*}
\mathrm{T\mbox{-}MAE}
=
\frac{1}{3N_aK}
\sum_{n=1}^{N_a}
\sum_{k=1}^{K}
\sum_{i=1}^{3}
\left|
a_{nk}^{(i)}-\hat{a}_{nk}^{(i)}
\right|,
\end{equation*}
\begin{equation*}
\mathrm{R\mbox{-}MAE}
=
\frac{1}{3N_aK}
\sum_{n=1}^{N_a}
\sum_{k=1}^{K}
\sum_{i=4}^{6}
\left|
a_{nk}^{(i)}-\hat{a}_{nk}^{(i)}
\right|,
\end{equation*}
where $\mathbf{a}_{nk}\in\mathbb{R}^6$ and $\hat{\mathbf{a}}_{nk}\in\mathbb{R}^6$ denote the ground-truth and predicted relative actions, respectively. The first three dimensions correspond to translation, and the last three correspond to Euler angles. Both metrics are averaged over the three corresponding dimensions and all ten target views. Detailed per-view results are provided in Sec.~\ref{sec:supp_perview_action}.

\subsection{Action-Conditioned World Modeling}
\label{sec:supp_world_model}

We evaluate whether the pretrained representations support action-conditioned prediction of future ultrasound observations. The world model consists of a Transformer predictor that predicts future latent features from the current observation and probe action, followed by a conditional diffusion decoder for pixel-level generation.

\paragraph{Predictor Architecture and Training Details.}

Given a current ultrasound observation $\mathbf{x}_t$ and a probe action $\mathbf{a}_t$, the pretrained encoder extracts its visual representation:
\begin{equation*}
\mathbf{z}_t=f_\theta(\mathbf{x}_t).
\end{equation*}
The probe action is first embedded through a linear layer and then provided together with the visual representation to a predictor, which estimates the future representation:
\begin{equation*}
\hat{\mathbf{z}}_{t+1}
=
P_\psi(\mathbf{z}_t,\mathbf{a}_t).
\end{equation*}

The predictor is trained to match the predicted future features with the features extracted from the corresponding target observation using Smooth L1 loss:
\begin{equation*}
\mathcal{L}_{\mathrm{WM}}
=
\operatorname{SmoothL1}
\left(
\hat{\mathbf{z}}_{t+1},
\mathbf{z}_{t+1}
\right),
\quad
\mathbf{z}_{t+1}=f_\theta(\mathbf{x}_{t+1}).
\end{equation*}

The predictor consists of four Transformer blocks with a hidden dimension of 384, six attention heads, and an MLP ratio of 2. It is trained for 30 epochs using AdamW with a batch size of 512 and a weight decay of 0.05. The learning rate starts at $10^{-4}$, with a five-epoch warm-up followed by cosine decay to $10^{-6}$. The architecture and training configurations are summarized in Table~\ref{tab:supp_world_predictor}.

\begin{table}[t]
    \centering
    \caption{\textbf{World model predictor configuration.}}
    \label{tab:supp_world_predictor}
    \vspace{-2mm}
    \small
    \begin{tabular}{lc}
        \toprule
        Configuration & Value \\
        \midrule
        \multicolumn{2}{c}{\textit{Architecture}} \\
        \midrule
        Transformer depth & 4 blocks \\
        Hidden dimension & 384 \\
        Attention heads & 6 \\
        MLP ratio & 2 \\
        Action embedding & Linear \\
        \midrule
        \multicolumn{2}{c}{\textit{Training}} \\
        \midrule
        Training epochs & 30 \\
        Batch size & 512 \\
        Optimizer & AdamW \\
        Initial learning rate & $10^{-4}$ \\
        Minimum learning rate & $10^{-6}$ \\
        Weight decay & 0.05 \\
        Warm-up & 5 epochs \\
        Learning rate schedule & Cosine decay \\
        Loss & Smooth L1 \\
        \bottomrule
    \end{tabular}
    \vspace{-5pt}
\end{table}

\paragraph{Diffusion Decoder Architecture and Training Details.}

To further evaluate the pixel-level generative capability of the world model, we train a generative model that maps the predicted features back to pixel space using the RCDM~\cite{bordes2021high} framework.
Specifically, we train an independent conditional diffusion decoder for each pretrained encoder. We first extract image features using the corresponding pretrained encoder and use them as conditioning information to reconstruct the original ultrasound images. During evaluation, the predicted future features are fed into the corresponding decoder to generate future observations:
\begin{equation*}
\hat{\mathbf{x}}_{t+1}
\sim
p_\omega
\left(
\mathbf{x}\mid\hat{\mathbf{z}}_{t+1}
\right).
\end{equation*}

All decoders adopt the same RCDM-style conditional U-Net architecture and training configuration to ensure a fair comparison. The output resolution is set to $128\times128$. The U-Net uses 128 base channels, channel multipliers of $(1,1,2,3,4)$, and two residual blocks per downsampling level. Attention modules are applied at resolutions of $32\times32$, $16\times16$, and $8\times8$, with four attention heads.

For training, we randomly sample 50,000 images from the training split and 2,048 images from the validation split. For each sampled image, we extract its visual features using the corresponding pretrained encoder and use them as conditioning inputs to train the diffusion decoder to reconstruct the original image. All target images are resized to $128\times128$. Each decoder is trained for a maximum of 100,000 optimizer steps using AdamW, with a batch size of 128 and a learning rate of $10^{-4}$. An exponential moving average (EMA) of the decoder parameters is maintained with a decay rate of 0.9999. The decoder architecture and training configurations are summarized in Table~\ref{tab:supp_diffusion_config}.

\begin{table}[t]
    \centering
    \caption{\textbf{World Model diffusion decoder configuration.}}
    \label{tab:supp_diffusion_config}
    \vspace{-2mm}
    \small
    \begin{tabular}{lc}
        \toprule
        Configuration & Value \\
        \midrule
        \multicolumn{2}{c}{\textit{Architecture}} \\
        \midrule
        Architecture & Conditional U-Net \\
        Output resolution & $128\times128$ \\
        Base channels & 128 \\
        Channel multipliers & $(1,1,2,3,4)$ \\
        Feature channels & $(128,128,256,384,512)$ \\
        Residual blocks per level & 2 \\
        Attention resolutions & $32,16,8$ \\
        Attention heads & 4 \\
        \midrule
        \multicolumn{2}{c}{\textit{Training}} \\
        \midrule
        Training images & 50,000 \\
        Validation images & 2,048 \\
        Maximum optimizer steps & 100,000 \\
        Batch size & 128 \\
        Optimizer & AdamW \\
        Learning rate & $10^{-4}$ \\
        EMA decay & 0.9999 \\
        \bottomrule
    \end{tabular}
    \vspace{-5pt}
\end{table}

\paragraph{Evaluation Protocol and Metrics.}

The world model is evaluated on the 65,520 test image pairs constructed from the 72 held-out scans. We evaluate predicted future observations at both the latent and pixel levels.

At the latent level, each predicted future representation is compared against the features of all images within the same scanning trajectory, and its nearest neighbor is retrieved. We compare the retrieved observation with the ground-truth target observation in terms of their relative 6D probe actions. Let $\mathbf{a}_n^{\mathrm{ret}}$ and $\mathbf{a}_n^{\mathrm{gt}}$ denote the corresponding retrieved and ground-truth actions, respectively. We report translation and rotation MAE (T-MAE and R-MAE), defined as:
\begin{equation*}
\mathrm{T\mbox{-}MAE}
=
\frac{1}{3N_w}
\sum_{n=1}^{N_w}
\sum_{i=1}^{3}
\left|
a_n^{\mathrm{ret},(i)}
-
a_n^{\mathrm{gt},(i)}
\right|,
\end{equation*}
\begin{equation*}
\mathrm{R\mbox{-}MAE}
=
\frac{1}{3N_w}
\sum_{n=1}^{N_w}
\sum_{i=4}^{6}
\left|
a_n^{\mathrm{ret},(i)}
-
a_n^{\mathrm{gt},(i)}
\right|,
\end{equation*}
where $N_w$ denotes the number of evaluated image pairs. The first three dimensions represent translation (mm), while the last three represent Euler angles ($^\circ$).

At the pixel level, the generated images and their corresponding ground-truth observations are both resized to $128\times128$ before evaluation. We report three standard image generation metrics: Learned Perceptual Image Patch Similarity (LPIPS), Fréchet Inception Distance (FID), and Structural Similarity Index Measure (SSIM). LPIPS measures perceptual differences between generated and ground-truth images, while FID evaluates the distributional similarity between generated and real images. SSIM measures their structural similarity. Lower LPIPS and FID indicate better generation quality, whereas higher SSIM indicates better structural consistency.

\subsection{Inverse Dynamics Prediction}
\label{sec:supp_inverse}

The inverse dynamics model predicts the relative probe movement between two ultrasound observations using their pretrained global representations.

\paragraph{Architecture.}

Given two observations $\mathbf{x}_i$ and $\mathbf{x}_j$, their representations are extracted using the shared encoder:
\begin{equation*}
\mathbf{z}_i=f_\theta(\mathbf{x}_i),
\quad
\mathbf{z}_j=f_\theta(\mathbf{x}_j).
\end{equation*}
The two representations are concatenated and fed into a two-layer MLP with an intermediate GELU activation to predict the 6D relative action:
\begin{equation*}
\hat{\mathbf{a}}_{i\rightarrow j}
=
g_\eta([\mathbf{z}_i;\mathbf{z}_j]).
\end{equation*}
The model is trained using Smooth L1 loss between the predicted and ground-truth relative actions:
\begin{equation*}
\mathcal{L}_{\mathrm{inv}}
=
\operatorname{SmoothL1}
\left(
\hat{\mathbf{a}}_{i\rightarrow j},
\mathbf{a}_{i\rightarrow j}
\right).
\end{equation*}

\paragraph{Training Details.}

We use AdamW to train the inverse dynamics for 5 epochs, with a batch size of 256, an initial learning rate of $10^{-4}$, and cosine learning rate scheduling.

\paragraph{Evaluation Protocol and Metrics.}

We evaluate inverse dynamics on the 65,520 test image pairs. We report translation and rotation mean absolute errors (T-MAE and R-MAE), computed over the 6D predicted and ground-truth actions:
\begin{equation*}
\mathrm{T\mbox{-}MAE}
=
\frac{1}{3N_p}
\sum_{n=1}^{N_p}
\sum_{i=1}^{3}
\left|
a_n^{(i)}-\hat{a}_n^{(i)}
\right|,
\end{equation*}
\begin{equation*}
\mathrm{R\mbox{-}MAE}
=
\frac{1}{3N_p}
\sum_{n=1}^{N_p}
\sum_{i=4}^{6}
\left|
a_n^{(i)}-\hat{a}_n^{(i)}
\right|,
\end{equation*}
where $\mathbf{a}_n\in\mathbb{R}^6$ and $\hat{\mathbf{a}}_n\in\mathbb{R}^6$ denote the ground-truth and predicted relative actions, respectively. The first three dimensions represent translation (mm), while the last three represent Euler angles ($^\circ$). Lower values indicate more accurate relative motion estimation.

\subsection{Latent Distance as a Progress Reward}
\label{sec:supp_reward}

Unlike the other downstream tasks, the reward model does not introduce an additional prediction head or require task-specific training. Given a current observation $\mathbf{x}_t$ and a target observation $\mathbf{x}_g$, we directly compute their latent distance:
\begin{equation*}
d^z(\mathbf{x}_t,\mathbf{x}_g)
=
\|f_\theta(\mathbf{x}_t)-f_\theta(\mathbf{x}_g)\|_2.
\end{equation*}
A smaller latent distance indicates greater proximity to the target. The same distance can therefore provide a progress signal for both human-guided and robotic scanning.

\paragraph{Evaluation Data and Metrics.}

We use the 65,520 test image pairs constructed from the held-out scans. For each pair, we calculate its physical distance and corresponding latent distance.
Then we evaluate the consistency between latent and physical distances using three metrics.

\textbf{Spearman's Rank Correlation.}
Spearman's $\rho$ measures the correlation between the ranks of physical and latent distances:
\begin{equation*}
\rho
=
\operatorname{Corr}
\left(
\operatorname{rank}(\{d_n^p\}),
\operatorname{rank}(\{d_n^z\})
\right).
\end{equation*}
Higher values indicate stronger monotonic agreement.

\textbf{Kendall's Rank Correlation.}
Kendall's $\tau$ measures the agreement between pairwise orderings. It is defined as:
\begin{equation*}
\tau
=
\frac{N_c-N_d}{N_c+N_d},
\end{equation*}
where $N_c$ and $N_d$ denote the numbers of concordant and discordant comparisons, respectively. Higher values indicate stronger ranking agreement.

\textbf{Pairwise Ranking Accuracy.}
Pairwise ranking accuracy measures the proportion of valid comparisons for which latent and physical distances have the same ordering:
\begin{equation*}
\mathrm{Acc}_{\mathrm{rank}}
=
\frac{1}{|\mathcal{Q}_{\mathrm{test}}|}
\sum_{(q,r)\in\mathcal{Q}_{\mathrm{test}}}
\mathbb{I}
\left[
(d_q^p-d_r^p)(d_q^z-d_r^z)>0
\right],
\end{equation*}
where $\mathcal{Q}_{\mathrm{test}}$ contains unordered comparisons with distinct physical distances, and $\mathbb{I}[\cdot]$ is the indicator function.

\section{Additional Evaluation Protocols}
\label{sec:supp_additional_eval}

\subsection{Probe Trajectory Reconstruction}
\label{sec:supp_trajectory}

Table~3 in the main paper reports the trajectory reconstruction errors. Here, we provide the details of the reconstruction procedure and evaluation metrics.

We construct 4,000 test trajectories from held-out test scans, each containing 100 temporally ordered ultrasound frames. For each pair of consecutive sampled observations, the inverse dynamics model predicts their relative probe transformation:
\begin{equation*}
\widehat{\Delta\mathbf{T}}_{t\rightarrow t+1}
=
\operatorname{InvDyn}
\left(
f_\theta(\mathbf{x}_t),
f_\theta(\mathbf{x}_{t+1})
\right).
\end{equation*}
Directly composing consecutive relative transformations provides an initial trajectory but accumulates prediction errors over time. To mitigate this drift and ensure a fair comparison, we apply the same sliding-window pose refinement strategy~\cite{Lategahn2012IV} to EchoACE and all baseline methods.

\paragraph{Sliding-Window Pose Refinement.}

The key idea is to use not only predictions between consecutive frames but also direct predictions between all frame pairs within a local window. For example, in a three-frame window (frames 0, 1, and 2), the model predicts the relative motions for all three pairs: 0$\rightarrow$1, 1$\rightarrow$2, and 0$\rightarrow$2. The direct prediction from frame 0 to frame 2 provides an additional constraint on the composition of the first two motions, helping reduce local drift.
At each step $k$, we construct a local window containing the current frame and up to five preceding sampled frames, indexed from $0$ to $m$, where $m=\min(5,k)$. We fix the first frame as the local reference, with $P_0=\mathbf{I}_{4\times4}$, and estimate the remaining poses $P_i=(R_i,\mathbf{t}_i)\in SE(3)$. Specifically, we first use the inverse dynamics model to predict relative motions between consecutive frames and compose these transformations sequentially to initialize the window poses. These initial estimates are then jointly refined using direct relative pose predictions between all frame pairs within the window.

For every frame pair $j<i$ within the window, the inverse dynamics model directly predicts the relative transformation from frame $j$ to frame $i$, following the same convention as the main paper:
\begin{equation*}
\widehat{\Delta\mathbf{T}}_{j\rightarrow i}
=
\begin{bmatrix}
\widehat{\Delta\mathbf{R}}_{ji} & \widehat{\Delta\mathbf{t}}_{ji}\\
\mathbf{0}^{\top} & 1
\end{bmatrix}.
\end{equation*}
Here, $\widehat{\Delta\mathbf{R}}_{ji}$ and $\widehat{\Delta\mathbf{t}}_{ji}$ denote the rotation and translation predicted directly by the inverse dynamics model. Meanwhile, the current estimated poses $P_i$ and $P_j$ imply another relative transformation, computed as $P_j^{-1}P_i$:
\begin{equation*}
P_j^{-1}P_i
=
\begin{bmatrix}
\Delta\mathbf{R}_{ji} & \Delta\mathbf{t}_{ji}\\
\mathbf{0}^{\top} & 1
\end{bmatrix},
\end{equation*}
where $\Delta\mathbf{R}_{ji}=R_j^\top R_i$ and $\Delta\mathbf{t}_{ji}=R_j^\top(\mathbf{t}_i-\mathbf{t}_j)$. Thus, the predicted transformation $(\widehat{\Delta\mathbf{R}}_{ji},\widehat{\Delta\mathbf{t}}_{ji})$ is compared with the relative transformation $(\Delta\mathbf{R}_{ji},\Delta\mathbf{t}_{ji})$ implied by the current pose estimates. We define their translation residual and rotation error as:
\begin{equation*}
\begin{aligned}
\mathbf{e}_{ji}
&=
\Delta\mathbf{t}_{ji}
-
\widehat{\Delta\mathbf{t}}_{ji},\\
\theta_{ji}
&=
\left\|
\left(
\operatorname{Log}
\left(
\widehat{\Delta\mathbf{R}}_{ji}^{\top}
\Delta\mathbf{R}_{ji}
\right)
\right)^\vee
\right\|_2.
\end{aligned}
\end{equation*}

We jointly refine all poses in the window by minimizing the discrepancies between the two sets of relative transformations:
\begin{equation*}
\begin{aligned}
\min_{P_1,\ldots,P_m\in SE(3)}
\quad &
\sum_{i=1}^{m}\sum_{j=0}^{i-1}
(i-j)^2
\left(
\frac{\|\mathbf{e}_{ji}\|_2^2}{\sigma_t^2}
+
\frac{\theta_{ji}^2}{\sigma_R^2}
\right),\\
&\text{s.t.}\quad P_0=\mathbf{I}_{4\times4}.
\end{aligned}
\end{equation*}
Here, $\sigma_t$ and $\sigma_R$ normalize the translation and rotation residuals, respectively. The weight $(i-j)^2$ increases the contribution of longer-range predictions according to their temporal separation. In particular, a prediction spanning five sampled steps receives 25 times the weight of an adjacent-frame prediction.

After optimization, we extract the refined relative motion between the last two frames and use it to update the global trajectory. Only the newly estimated pose is appended, while previously reconstructed poses remain fixed:
\begin{equation*}
\Delta_k=(P_{m-1}^*)^{-1}P_m^*,
\quad
G_k=G_{k-1}\Delta_k,
\quad
G_0=\mathbf{I}_{4\times4},
\end{equation*}
where $P_i^*$ denotes the optimized local pose and $G_k$ denotes the accumulated global pose. The window then advances by one sampled frame, while previously reconstructed global poses remain fixed. This procedure relies only on inverse dynamics predictions and requires no ground-truth poses during refinement.

\paragraph{Implementation Details.}

We subsample each original scanning sequence by retaining every 25th available frame. Each evaluation trajectory consists of 100 temporally ordered sampled frames. A full refinement window therefore contains six sampled frames and 15 pairwise constraints. We set the translation and rotation normalization scales to $\sigma_t=1$ mm and $\sigma_R=1^\circ=\pi/180$ radians, respectively. The nonlinear least-squares problem is solved using the trust-region reflective algorithm (\texttt{trf}), with a maximum of 60 function evaluations per window. The convergence tolerances \texttt{ftol}, \texttt{xtol}, and \texttt{gtol} are all set to $10^{-7}$.

\paragraph{Evaluation Metrics.}

We evaluate the reconstructed trajectories using Position RMSE and Mean Rotation Error, measuring translational and rotational discrepancies, respectively. Let $N_{\mathrm{traj}}=4000$ denote the number of test trajectories and $T=100$ the number of frames per trajectory. For each trajectory, both the ground-truth and reconstructed poses are expressed relative to the initial frame, which serves as the reference coordinate system.

Position RMSE measures the average Euclidean position error:
\begin{equation*}
\mathrm{Position\ RMSE}
=
\sqrt{
\frac{1}{N_{\mathrm{traj}}T}
\sum_{n=1}^{N_{\mathrm{traj}}}
\sum_{t=1}^{T}
\left\|
\hat{\mathbf{p}}_{n,t}
-
\mathbf{p}_{n,t}
\right\|_2^2
}.
\end{equation*}

Mean Rotation Error measures the average geodesic angular discrepancy between reconstructed and ground-truth orientations. For each frame, the angular error in radians is defined as:
\begin{equation*}
\theta_{n,t}
=
\left\|
\operatorname{Log}
\left(
\mathbf{R}_{n,t}^{\top}
\hat{\mathbf{R}}_{n,t}
\right)^\vee
\right\|_2.
\end{equation*}
The Mean Rotation Error in degrees is computed as:
\begin{equation*}
\mathrm{Mean\ Rotation\ Error}
=
\frac{180}{\pi N_{\mathrm{traj}}T}
\sum_{n=1}^{N_{\mathrm{traj}}}
\sum_{t=1}^{T}
\theta_{n,t}.
\end{equation*}
Here, $\mathbf{p}_{n,t}$ and $\hat{\mathbf{p}}_{n,t}$ denote the ground-truth and reconstructed positions, respectively, while $\mathbf{R}_{n,t}$ and $\hat{\mathbf{R}}_{n,t}$ denote their corresponding rotation matrices. Lower values indicate more accurate trajectory reconstruction.

\subsection{Left-Ventricular 3D Reconstruction}
\label{sec:supp_3d}

Figure~\ref{fig:recon}(b) in the main paper visualizes the 3D left-ventricular reconstructions obtained using EchoACE and baseline methods. Here, we detail the implementation of this experiment. We select ten end-diastolic ultrasound images from a single case, acquired at different probe poses, in which the left ventricle is clearly visible. We segment the left-ventricular boundaries to obtain corresponding 2D binary masks. All methods use the same images, segmentation masks, and reconstruction pipeline, differing only in the probe poses estimated by their respective inverse dynamics models (EchoACE, UltraSeP, and EchoCLIP).

\textbf{Coordinate Transformation and Scale.}
Image pixels are converted into physical coordinates using the ultrasound image scale. The corresponding inverse-dynamics-estimated probe poses and an estimated image-to-probe transformation are then used to place each 2D mask in a common 3D coordinate system.

\textbf{3D Fusion and Interpolation.}
Each 2D mask is converted into a signed distance map, where positive values indicate locations inside the cavity, negative values indicate locations outside, and zero corresponds to the boundary. We construct a 3D grid with a voxel spacing of 1.5 mm. For each voxel, its position is projected onto each ultrasound image plane, and the corresponding signed distance is obtained through bilinear interpolation. Contributions from image planes farther from the voxel are reduced using Gaussian weights based on the voxel-to-plane distance, with a standard deviation of 10 mm. The weighted distance values from the ten views are then combined, and voxels with positive fused distances and sufficient multi-view support are classified as part of the cavity.

\textbf{Volume and Surface Post-processing.}
The reconstructed binary cavity is processed using morphological closing to bridge small gaps, followed by hole filling and retention of the largest connected component. Gaussian filtering is then applied to reduce voxel-level irregularities. Finally, the cavity boundary is extracted as a triangular mesh using marching cubes, followed by light Laplacian smoothing. The final 3D model is exported in VTP format.

\subsection{Progress Reward Baselines in Figure 9}
\label{sec:supp_reward_baselines}

Figure~\ref{fig:vis_reward}(b,c) in the main paper shows the progress estimation results using a standard-view classification model and a vision-language model (VLM) as reward signals, respectively. We describe their implementations below.

\paragraph{View Classification Confidence.}

We train a ten-class standard-view classification model on the training split. The model uses ResNet-34 as the image encoder, followed by a three-layer MLP with GELU activations between consecutive layers. It outputs classification probabilities over the ten standard cardiac views. Given a target view, we directly use the predicted confidence of its corresponding class, ranging from 0 to 1, as the progress reward.

\paragraph{VLM-Generated Reward.}

We use ChatGPT 5.6 Sol as the VLM. Given the current ultrasound image and the target view, the model is prompted to directly assign a progress reward score between 0 and 1, indicating the current scanning progress toward the target view. The generated score is used directly as the reward signal.

\section{Additional Results}
\label{sec:supp_results}

\begin{figure}[t]
    \centering
    \includegraphics[width=\linewidth]{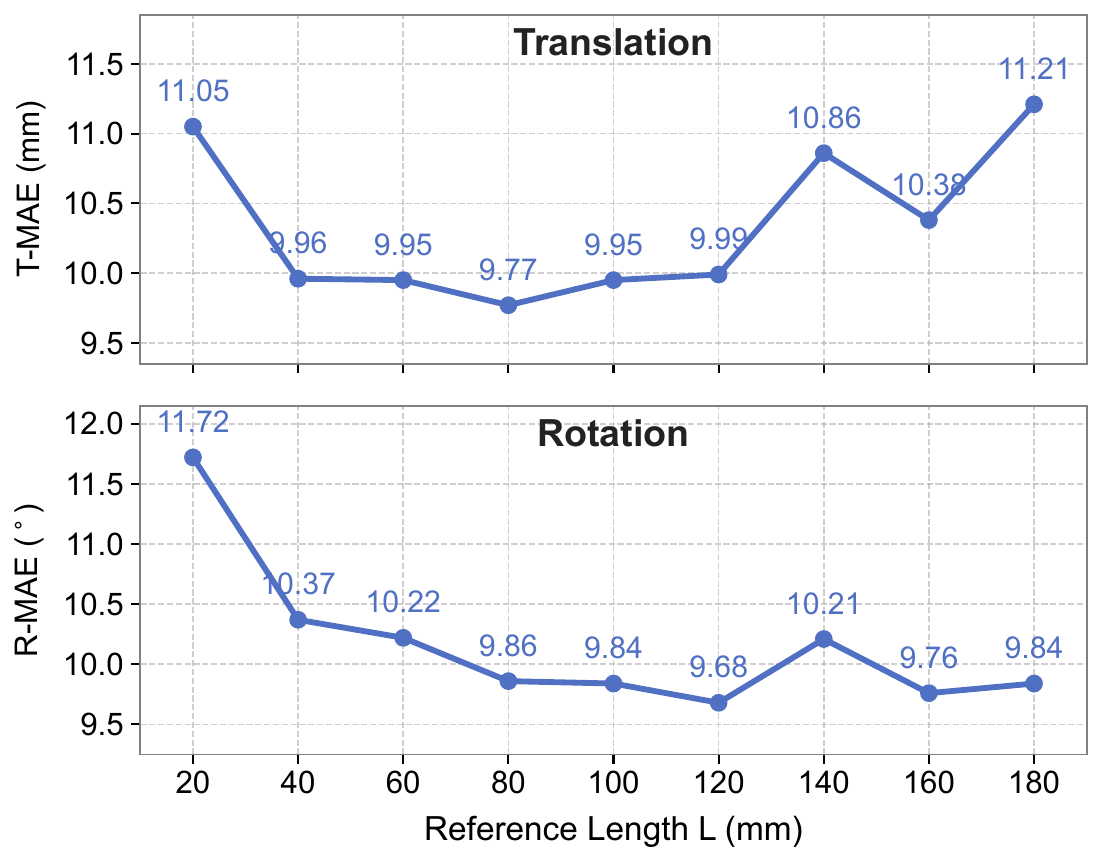}
    \caption{
        \textbf{Sensitivity to the reference length $L$.}
        Translation and rotation MAE of the single-frame Action Model under different reference lengths.
    }
    \label{fig:reference_length_sensitivity}
    \vspace{-10pt}
\end{figure}

\subsection{Sensitivity to the Reference Length}
\label{sec:supp_reference_length}

In the main paper, we introduce a reference length $L$ to convert rotational displacement into an equivalent linear displacement, serving as an effective lever arm from the probe's rotation center. This parameter determines the relative contribution of translation and rotation to the physical distance used for ranking supervision. We set $L=100\,\mathrm{mm}$ by default based on typical probe-motion scales. To investigate the sensitivity of EchoACE to this choice, we evaluate different reference lengths ranging from 20 to 180 mm.

Specifically, we vary $L$ while keeping all other pretraining configurations unchanged. For each setting, we evaluate the resulting encoder using the single-frame Action Model and report translation and rotation MAE (T-MAE and R-MAE), averaged over the ten standard cardiac views.
As shown in Fig.~\ref{fig:reference_length_sensitivity}, both metrics exhibit relatively stable performance around $L=100\,\mathrm{mm}$. The lowest translation error is achieved at $L=80\,\mathrm{mm}$ (9.77 mm), while the lowest rotation error occurs at $L=120\,\mathrm{mm}$ (9.68$^\circ$). At $L=100\,\mathrm{mm}$, the corresponding errors are 9.95 mm and 9.84$^\circ$, respectively, remaining close to their minimum values. In contrast, substantially smaller or larger values can lead to increased prediction errors, particularly in translation. These results indicate that EchoACE is relatively insensitive to the reference length within a reasonable range, supporting the use of $L=100\,\mathrm{mm}$ as a robust default choice.

\subsection{Per-View Action Prediction Errors}
\label{sec:supp_perview_action}

The action prediction errors reported in the main paper are averaged over the ten standard views. Here, we provide the translation and rotation MAE for each individual view. Specifically, the per-view results for the single-frame and sequence Action Models in Table~1 are reported in Tables~\ref{tab:single_per_view} and~\ref{tab:seq_per_view}, respectively. The per-view results for the fine-tuning and ablation experiments in Tables~4 and~6 are provided in Tables~\ref{tab:tuning_per_view} and~\ref{tab:ablation_per_view}, respectively.

\subsection{Cross-Device External Validation}
\label{sec:supp_external_val}

To further evaluate model performance under cross-device variations, we conduct an independent external test using data collected with a different ultrasound system and probe. Specifically, we use a Siemens ultrasound system equipped with a 4V1c probe to acquire data from 20 new subjects, forming an independent external test set.

\paragraph{External Test Set.}
The original test set used in the main experiments was collected using a GE ultrasound system with an M5S probe, whereas the external test set was acquired using a Siemens system with a 4V1c probe. As summarized in Table~\ref{tab:supp_external_dataset}, the original test set contains 36 subjects, 72 scans, and 291,954 images, while the external test set contains 20 subjects, 20 scans, and 186,606 images.

\paragraph{Evaluation Protocol and Results.}

We evaluate the single-frame Action Model using the pretrained weights from the original GE dataset. We directly apply the previously trained models to the external test set without additional fine-tuning or adaptation. We compare EchoACE with UltraSeP and report translation and rotation MAE (T-MAE and R-MAE), averaged over the ten standard views.

As shown in Fig.~\ref{fig:supp_external_results}, both methods exhibit some performance degradation on the external test set, indicating that differences in ultrasound systems and probes introduce additional challenges. Nevertheless, EchoACE consistently outperforms UltraSeP in both translation and rotation, maintaining a clear performance advantage despite the cross-device domain shift.

\begin{table*}[t]
    \centering

    % ========================================================
    % LEFT: External Dataset Comparison
    % ========================================================
    \begin{minipage}[t]{0.47\textwidth}
        \vspace{0pt}
        \vspace*{9mm}
        \centering

        \caption{
            \textbf{Comparison of the original and external test sets.}
        }
        \label{tab:supp_external_dataset}

        \small
        \setlength{\tabcolsep}{5pt}
        \renewcommand{\arraystretch}{1.3}
        \begin{tabular}{lcc}
            \toprule
            & \shortstack{Original Test\\Set}
            & \shortstack{External Test\\Set} \\
            \midrule
            Ultrasound System & GE & Siemens \\
            Probe & M5S & 4V1c \\
            Subjects & 36 & 20 \\
            Scans & 72 & 20 \\
            Images & 291,954 & 186,606 \\
            \bottomrule
        \end{tabular}
    \end{minipage}
    \hfill
    % ========================================================
    % RIGHT: External Validation Results
    % ========================================================
    \begin{minipage}[t]{0.47\textwidth}
        \vspace{0pt}
        \centering

        \includegraphics[width=\linewidth]{
            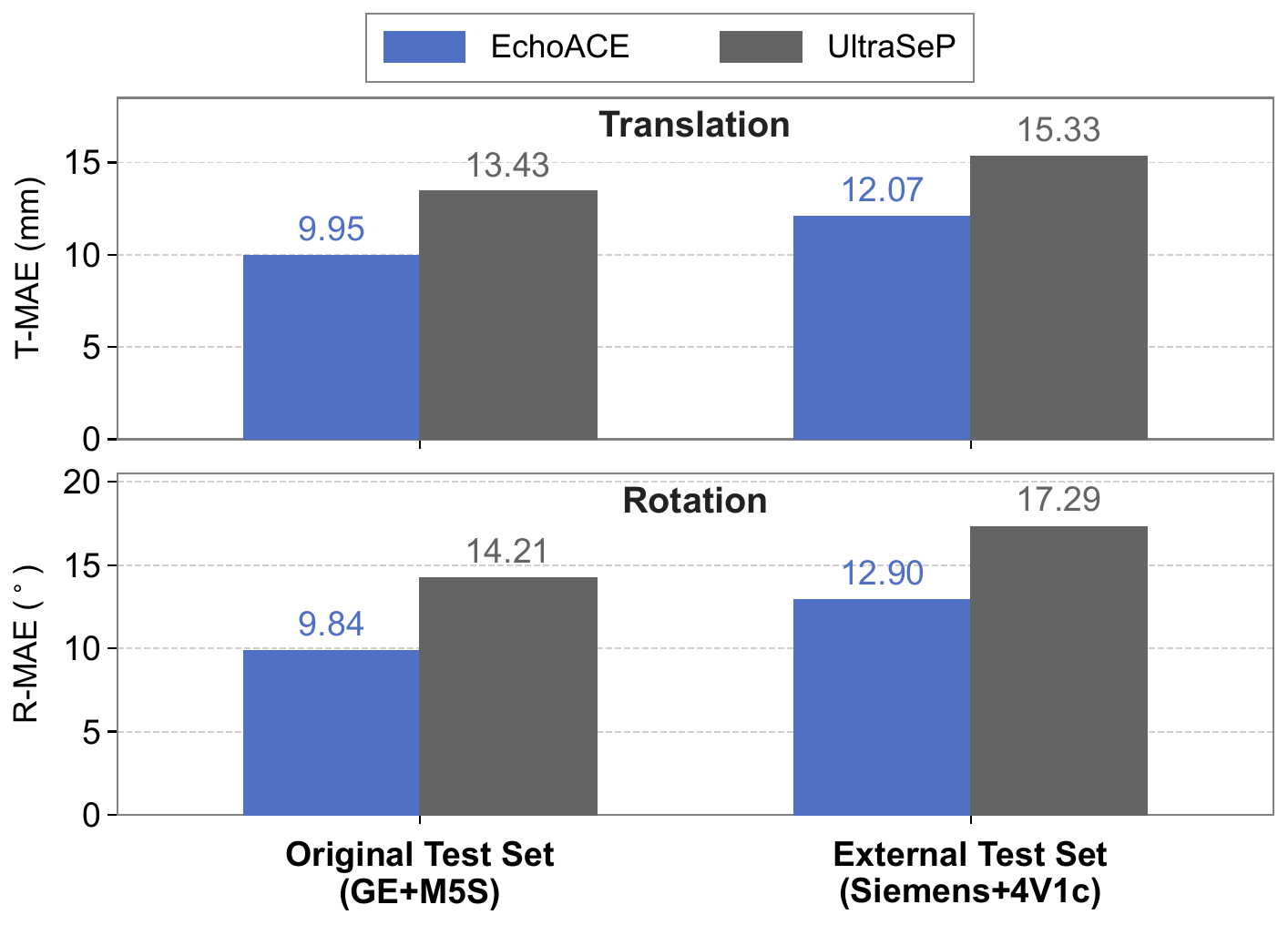
        }

        \captionof{figure}{
            \textbf{Cross-device external test results.}
            Comparison of EchoACE and UltraSeP on the original GE test set and the external Siemens test set, reporting translation and rotation MAE.
        }
        \label{fig:supp_external_results}
    \end{minipage}

    \vspace{-3mm}
\end{table*}

\newcolumntype{Q}{
    >{\raggedright\arraybackslash}m{2.35cm}
}

\newcolumntype{Y}{
    >{\centering\arraybackslash}X
}

\begin{table*}[t]
    \centering

    \caption{
        \textbf{Per-view errors} of the \textbf{single-frame Action Model}
        in the main paper's \textbf{Table~\ref{tab:result_main}} across ten standard
        views.
        Translation and rotation errors (Trans./Rot.) are reported
        as mean absolute errors in millimeters and degrees,
        respectively.
    }

    \label{tab:single_per_view}

    \vspace{-2mm}

    \begingroup

    \footnotesize

    \setlength{\tabcolsep}{0.6pt}

    \renewcommand{\arraystretch}{1.15}

    \renewcommand{\tabularxcolumn}[1]{m{#1}}

    \begin{tabularx}{\textwidth}{
        @{}
        Q
        *{20}{Y}
        @{}
    }

        \toprule

        \multirow[c]{2}{*}{
            \footnotesize\textbf{Method}
        }
        &
        \multicolumn{2}{c}{
            \scriptsize\bfseries PLAX
        }
        &
        \multicolumn{2}{c}{
            \scriptsize\bfseries PSAX-AV
        }
        &
        \multicolumn{2}{c}{
            \scriptsize\bfseries PSAX-PV
        }
        &
        \multicolumn{2}{c}{
            \scriptsize\bfseries PSAX-MV
        }
        &
        \multicolumn{2}{c}{
            \scriptsize\bfseries PSAX-PAP
        }
        &
        \multicolumn{2}{c}{
            \scriptsize\bfseries PSAX-APEX
        }
        &
        \multicolumn{2}{c}{
            \scriptsize\bfseries A4C
        }
        &
        \multicolumn{2}{c}{
            \scriptsize\bfseries A5C
        }
        &
        \multicolumn{2}{c}{
            \scriptsize\bfseries A3C
        }
        &
        \multicolumn{2}{c}{
            \scriptsize\bfseries A2C
        }
        \\

        \cmidrule(lr){2-3}
        \cmidrule(lr){4-5}
        \cmidrule(lr){6-7}
        \cmidrule(lr){8-9}
        \cmidrule(lr){10-11}
        \cmidrule(lr){12-13}
        \cmidrule(lr){14-15}
        \cmidrule(lr){16-17}
        \cmidrule(lr){18-19}
        \cmidrule(lr){20-21}

        &
        \scriptsize Trans. & \scriptsize Rot.
        &
        \scriptsize Trans. & \scriptsize Rot.
        &
        \scriptsize Trans. & \scriptsize Rot.
        &
        \scriptsize Trans. & \scriptsize Rot.
        &
        \scriptsize Trans. & \scriptsize Rot.
        &
        \scriptsize Trans. & \scriptsize Rot.
        &
        \scriptsize Trans. & \scriptsize Rot.
        &
        \scriptsize Trans. & \scriptsize Rot.
        &
        \scriptsize Trans. & \scriptsize Rot.
        &
        \scriptsize Trans. & \scriptsize Rot.
        \\

        \midrule

        \multicolumn{21}{l}{
            \footnotesize\itshape
            Appearance-Centric Representations,
            Generic Pretrained Vision Foundation Models
        }
        \\

        \addlinespace[2pt]

        SigLIP 2~\cite{tschannen2025siglip}
        & 17.95 & 20.14
        & 16.65 & 19.17
        & 18.33 & 20.25
        & 17.10 & 20.60
        & 16.14 & 20.71
        & 16.38 & 21.43
        & 20.20 & 21.62
        & 20.95 & 22.61
        & 21.36 & 22.09
        & 20.78 & 22.75
        \\

        DINOv3~\cite{simeoni2025dinov3}
        & 15.82 & 18.52
        & 14.53 & 18.01
        & 15.94 & 18.55
        & 14.81 & 19.37
        & 14.20 & 19.42
        & 15.45 & 19.82
        & 19.02 & 19.86
        & 19.73 & 21.01
        & 19.99 & 20.81
        & 19.55 & 21.36
        \\

        \midrule

        \multicolumn{21}{l}{
            \footnotesize\itshape
            Appearance-Centric Representations,
            Pre-trained on Our Ultrasound Data
        }
        \\

        \addlinespace[2pt]

        US-MAE~\cite{mae}
        & 16.30 & 17.90
        & 15.69 & 17.77
        & 17.03 & 18.58
        & 15.92 & 19.13
        & 15.15 & 19.17
        & 15.83 & 19.57
        & 19.32 & 19.35
        & 19.84 & 20.55
        & 19.99 & 20.27
        & 19.70 & 21.10
        \\

        US-MoCo~\cite{mocov3}
        & 15.59 & 18.11
        & 14.80 & 17.83
        & 15.92 & 18.50
        & 14.88 & 19.17
        & 14.13 & 19.20
        & 15.49 & 19.72
        & 18.71 & 19.67
        & 19.44 & 20.60
        & 19.38 & 20.07
        & 19.57 & 21.07
        \\

        \midrule

        \multicolumn{21}{l}{
            \footnotesize\itshape
            Appearance-Centric Representations,
            Open-source Medical and Ultrasound Foundation Models
        }
        \\

        \addlinespace[2pt]

        USFM~\cite{jiao2024usfm}
        & 17.97 & 20.62
        & 16.10 & 18.80
        & 17.97 & 20.27
        & 16.36 & 20.06
        & 15.50 & 20.26
        & 16.13 & 21.04
        & 20.47 & 21.88
        & 21.37 & 23.14
        & 22.01 & 22.97
        & 20.82 & 22.77
        \\

        LVM-Med~\cite{mh2024lvm}
        & 15.53 & 17.67
        & 14.51 & 17.60
        & 15.88 & 18.16
        & 14.86 & 19.06
        & 14.05 & 18.98
        & 15.14 & 19.46
        & 19.19 & 19.19
        & 19.93 & 20.08
        & 19.91 & 20.14
        & 19.50 & 20.86
        \\

        BioMedCLIP~\cite{zhang2023biomedclip}
        & 14.53 & 18.39
        & 13.84 & 17.84
        & 15.07 & 18.26
        & 14.35 & 18.88
        & 13.35 & 19.14
        & 14.47 & 19.54
        & 18.54 & 19.48
        & 19.30 & 20.22
        & 19.25 & 20.11
        & 18.60 & 21.11
        \\

        EchoCLIP~\cite{christensen2024vision}
        & 13.54 & 16.21
        & 11.99 & 15.91
        & 13.32 & 16.18
        & \secondbest{12.30} & 17.33
        & 11.72 & 17.38
        & 13.64 & 17.99
        & 16.86 & 16.94
        & 17.57 & 18.05
        & 17.74 & 18.52
        & 17.19 & 19.36
        \\

        \midrule

        \multicolumn{21}{l}{
            \footnotesize\itshape
            Action-Conditioned Representations,
            Pre-trained on Our Ultrasound Data
        }
        \\

        \addlinespace[2pt]

        Structure-aware\cite{jiang2024structure}
        & 15.54 & 17.07
        & 14.69 & 16.71
        & 15.86 & 17.35
        & 15.10 & 18.35
        & 14.42 & 18.39
        & 15.40 & 18.60
        & 18.68 & 18.47
        & 19.46 & 19.52
        & 19.40 & 19.37
        & 19.23 & 20.38
        \\

        EchoWorld~\cite{yue2025echoworld}
        & 15.50 & 17.10
        & 14.52 & 16.73
        & 15.87 & 17.30
        & 14.87 & 18.43
        & 14.25 & 18.52
        & 15.29 & 18.78
        & 18.92 & 18.34
        & 19.58 & 19.34
        & 19.52 & 19.77
        & 19.18 & 20.59
        \\

        UltraSeP~\cite{jiang2025ultrasep}
        & \secondbest{13.16} & \secondbest{12.99}
        & \secondbest{11.62} & \secondbest{12.35}
        & \secondbest{12.96} & \secondbest{12.92}
        & 12.31 & \secondbest{13.80}
        & \secondbest{11.62} & \secondbest{13.69}
        & \secondbest{12.29} & \secondbest{14.01}
        & \secondbest{14.83} & \secondbest{14.14}
        & \secondbest{16.04} & \secondbest{16.20}
        & \secondbest{14.54} & \secondbest{15.13}
        & \secondbest{14.94} & \secondbest{16.86}
        \\

        \midrule

        \multicolumn{21}{l}{
            \footnotesize\itshape
            Action-Centric Representations,
            Pre-trained on Our Ultrasound Data
        }
        \\

        \addlinespace[2pt]

        \textbf{EchoACE (Ours)}
        & \best{10.24} & \best{8.64}
        & \best{9.10} & \best{8.66}
        & \best{9.73} & \best{9.01}
        & \best{9.29} & \best{9.45}
        & \best{8.88} & \best{9.62}
        & \best{10.04} & \best{9.99}
        & \best{10.44} & \best{8.66}
        & \best{11.14} & \best{10.46}
        & \best{10.21} & \best{11.31}
        & \best{10.42} & \best{12.59}
        \\

        \bottomrule

    \end{tabularx}

    \endgroup

    \vspace{-5pt}

\end{table*}

\newcolumntype{Q}{
    >{\raggedright\arraybackslash}m{2.35cm}
}

\newcolumntype{Y}{
    >{\centering\arraybackslash}X
}

\begin{table*}[t]
    \centering

    \caption{
        \textbf{Per-view errors} of the \textbf{sequence Action Model}
        in the main paper's \textbf{Table~\ref{tab:result_main}} across ten standard
        views.
        Translation and rotation errors (Trans./Rot.) are reported
        as mean absolute errors in millimeters and degrees,
        respectively.
    }

    \label{tab:seq_per_view}

    \vspace{-2mm}

    \begingroup

    \footnotesize

    \setlength{\tabcolsep}{0.6pt}

    \renewcommand{\arraystretch}{1.15}

    \renewcommand{\tabularxcolumn}[1]{m{#1}}

    \begin{tabularx}{\textwidth}{
        @{}
        Q
        *{20}{Y}
        @{}
    }

        \toprule

        \multirow[c]{2}{*}{
            \footnotesize\textbf{Method}
        }
        &
        \multicolumn{2}{c}{
            \scriptsize\bfseries PLAX
        }
        &
        \multicolumn{2}{c}{
            \scriptsize\bfseries PSAX-AV
        }
        &
        \multicolumn{2}{c}{
            \scriptsize\bfseries PSAX-PV
        }
        &
        \multicolumn{2}{c}{
            \scriptsize\bfseries PSAX-MV
        }
        &
        \multicolumn{2}{c}{
            \scriptsize\bfseries PSAX-PAP
        }
        &
        \multicolumn{2}{c}{
            \scriptsize\bfseries PSAX-APEX
        }
        &
        \multicolumn{2}{c}{
            \scriptsize\bfseries A4C
        }
        &
        \multicolumn{2}{c}{
            \scriptsize\bfseries A5C
        }
        &
        \multicolumn{2}{c}{
            \scriptsize\bfseries A3C
        }
        &
        \multicolumn{2}{c}{
            \scriptsize\bfseries A2C
        }
        \\

        \cmidrule(lr){2-3}
        \cmidrule(lr){4-5}
        \cmidrule(lr){6-7}
        \cmidrule(lr){8-9}
        \cmidrule(lr){10-11}
        \cmidrule(lr){12-13}
        \cmidrule(lr){14-15}
        \cmidrule(lr){16-17}
        \cmidrule(lr){18-19}
        \cmidrule(lr){20-21}

        &
        \scriptsize Trans. & \scriptsize Rot.
        &
        \scriptsize Trans. & \scriptsize Rot.
        &
        \scriptsize Trans. & \scriptsize Rot.
        &
        \scriptsize Trans. & \scriptsize Rot.
        &
        \scriptsize Trans. & \scriptsize Rot.
        &
        \scriptsize Trans. & \scriptsize Rot.
        &
        \scriptsize Trans. & \scriptsize Rot.
        &
        \scriptsize Trans. & \scriptsize Rot.
        &
        \scriptsize Trans. & \scriptsize Rot.
        &
        \scriptsize Trans. & \scriptsize Rot.
        \\

        \midrule

        \multicolumn{21}{l}{
            \footnotesize\itshape
            Appearance-Centric Representations,
            Generic Pretrained Vision Foundation Models
        }
        \\

        \addlinespace[2pt]

        SigLIP 2~\cite{tschannen2025siglip}
        & 8.87 & 7.31
        & 8.43 & 7.96
        & 8.99 & 8.02
        & 8.27 & 8.47
        & 8.10 & 8.34
        & 9.64 & 9.42
        & 8.53 & 7.73
        & 8.89 & 9.68
        & 8.52 & 9.43
        & 8.52 & 11.28
        \\

        DINOv3~\cite{simeoni2025dinov3}
        & 8.44 & 7.13
        & 7.82 & 7.74
        & 8.30 & 7.45
        & 7.74 & 8.50
        & 7.44 & 8.34
        & 8.87 & 9.20
        & 8.22 & 7.42
        & 8.49 & 9.35
        & 8.49 & 9.21
        & 8.35 & 11.18
        \\

        \midrule

        \multicolumn{21}{l}{
            \footnotesize\itshape
            Appearance-Centric Representations,
            Pre-trained on Our Ultrasound Data
        }
        \\

        \addlinespace[2pt]

        US-MAE~\cite{mae}
        & 8.13 & 6.94
        & 7.83 & 7.69
        & 8.29 & 7.50
        & 7.58 & 8.29
        & 7.45 & 8.15
        & 8.93 & 9.00
        & 8.19 & 7.31
        & 8.45 & 9.28
        & 8.44 & 9.16
        & 8.34 & 11.26
        \\

        US-MoCo~\cite{mocov3}
        & 8.43 & 7.08
        & 7.99 & 7.61
        & 8.57 & 7.52
        & 7.69 & 8.41
        & 7.54 & 8.33
        & 8.75 & 9.20
        & 8.19 & 7.34
        & 8.64 & 9.35
        & 8.54 & 9.29
        & 8.31 & 11.45
        \\

        \midrule

        \multicolumn{21}{l}{
            \footnotesize\itshape
            Appearance-Centric Representations,
            Open-source Medical and Ultrasound Foundation Models
        }
        \\

        \addlinespace[2pt]

        USFM~\cite{jiao2024usfm}
        & 8.49 & 6.90
        & 7.79 & 7.56
        & 8.21 & 7.46
        & 7.71 & 8.12
        & 7.43 & 8.09
        & 8.84 & 8.83
        & 8.19 & 7.42
        & 8.58 & 9.61
        & 8.62 & 9.59
        & 8.42 & 11.47
        \\

        LVM-Med~\cite{mh2024lvm}
        & 8.27 & 6.91
        & 7.75 & 7.51
        & 8.17 & 7.47
        & 7.49 & 8.44
        & 7.29 & 8.23
        & 8.73 & 8.99
        & 8.23 & 7.46
        & 8.56 & 9.63
        & 8.40 & 9.35
        & 8.35 & 11.55
        \\

        BioMedCLIP~\cite{zhang2023biomedclip}
        & 8.18 & 7.05
        & 7.84 & 7.67
        & 8.23 & 7.62
        & 7.58 & 8.24
        & 7.38 & 8.01
        & 8.66 & 8.91
        & 8.01 & 7.38
        & 8.19 & 9.38
        & 8.18 & 8.93
        & 8.14 & 11.13
        \\

        EchoCLIP~\cite{christensen2024vision}
        & \secondbest{7.70} & 6.59
        & 7.46 & 7.47
        & 7.87 & 7.27
        & 7.20 & 8.05
        & 7.06 & 7.87
        & 8.55 & 8.74
        & 7.77 & 7.10
        & 7.95 & 9.21
        & 8.10 & 8.80
        & 7.86 & 10.85
        \\

        \midrule

        \multicolumn{21}{l}{
            \footnotesize\itshape
            Action-Conditioned Representations,
            Pre-trained on Our Ultrasound Data
        }
        \\

        \addlinespace[2pt]

        Structure-aware\cite{jiang2024structure}
        & 7.97 & 6.82
        & 7.42 & 7.47
        & 7.94 & 7.34
        & 7.21 & 8.04
        & 7.09 & 8.07
        & 8.67 & 8.64
        & 7.79 & 7.07
        & 8.15 & 9.22
        & 8.44 & 9.30
        & 8.01 & 11.12
        \\

        EchoWorld~\cite{yue2025echoworld}
        & 7.90 & 6.65
        & 7.65 & 7.45
        & 8.00 & 7.37
        & 7.25 & 8.09
        & 7.14 & 7.96
        & 8.63 & 8.62
        & 7.95 & 7.13
        & 8.26 & 9.31
        & 8.36 & 9.04
        & 8.09 & 11.25
        \\

        UltraSeP~\cite{jiang2025ultrasep}
        & 7.83 & \secondbest{6.43}
        & \secondbest{7.33} & \secondbest{7.11}
        & \secondbest{7.77} & \secondbest{7.18}
        & \secondbest{7.01} & \secondbest{7.79}
        & \secondbest{6.98} & \secondbest{7.82}
        & \secondbest{8.51} & \secondbest{8.43}
        & \secondbest{7.53} & \secondbest{6.64}
        & \secondbest{7.75} & \secondbest{8.81}
        & \secondbest{8.00} & \secondbest{8.71}
        & \secondbest{7.68} & \secondbest{10.69}
        \\

        \midrule

        \multicolumn{21}{l}{
            \footnotesize\itshape
            Action-Centric Representations,
            Pre-trained on Our Ultrasound Data
        }
        \\

        \addlinespace[2pt]

        \textbf{EchoACE (Ours)}
        & \best{6.73} & \best{5.82}
        & \best{6.85} & \best{6.64}
        & \best{7.45} & \best{6.73}
        & \best{6.43} & \best{7.19}
        & \best{6.39} & \best{6.97}
        & \best{7.80} & \best{7.54}
        & \best{7.01} & \best{6.16}
        & \best{7.05} & \best{8.56}
        & \best{7.35} & \best{8.08}
        & \best{7.18} & \best{10.21}
        \\

        \bottomrule

    \end{tabularx}

    \endgroup

    \vspace{-5pt}

\end{table*}

% Method column
\newcolumntype{Q}{
    >{\raggedright\arraybackslash}m{2.35cm}
}

% Centered expandable columns for numerical results
\newcolumntype{Y}{
    >{\centering\arraybackslash}X
}

\begin{table*}[t]
    \centering

    \caption{
        \textbf{Per-view errors} of the \textbf{fine-tuned single-frame Action Model} in the main paper's \textbf{Table~\ref{tab:finetuning_results}} across ten standard
        views.
        Translation and rotation errors (Trans./Rot.) are reported
        as mean absolute errors in millimeters and degrees,
        respectively.
    }

    \label{tab:tuning_per_view}

    \vspace{-2mm}

    \begingroup

    \footnotesize

    % Horizontal spacing
    \setlength{\tabcolsep}{0.6pt}

    % Vertical spacing
    \renewcommand{\arraystretch}{1.15}

    % Vertically center all X columns
    \renewcommand{\tabularxcolumn}[1]{m{#1}}

    % ========================================================
    % 21 columns:
    % 1: Method
    % 2-21: 10 views, each with Trans. and Rot.
    % ========================================================

    \begin{tabularx}{\textwidth}{
        @{}
        Q
        *{20}{Y}
        @{}
    }

        \toprule

        % ==========================================
        % First header row: Echocardiographic views
        % ==========================================

        \multirow[c]{2}{*}{
            \footnotesize\textbf{Method}
        }
        &
        \multicolumn{2}{c}{
            \scriptsize\bfseries PLAX
        }
        &
        \multicolumn{2}{c}{
            \scriptsize\bfseries PSAX-AV
        }
        &
        \multicolumn{2}{c}{
            \scriptsize\bfseries PSAX-PV
        }
        &
        \multicolumn{2}{c}{
            \scriptsize\bfseries PSAX-MV
        }
        &
        \multicolumn{2}{c}{
            \scriptsize\bfseries PSAX-PAP
        }
        &
        \multicolumn{2}{c}{
            \scriptsize\bfseries PSAX-APEX
        }
        &
        \multicolumn{2}{c}{
            \scriptsize\bfseries A4C
        }
        &
        \multicolumn{2}{c}{
            \scriptsize\bfseries A5C
        }
        &
        \multicolumn{2}{c}{
            \scriptsize\bfseries A3C
        }
        &
        \multicolumn{2}{c}{
            \scriptsize\bfseries A2C
        }
        \\

        \cmidrule(lr){2-3}
        \cmidrule(lr){4-5}
        \cmidrule(lr){6-7}
        \cmidrule(lr){8-9}
        \cmidrule(lr){10-11}
        \cmidrule(lr){12-13}
        \cmidrule(lr){14-15}
        \cmidrule(lr){16-17}
        \cmidrule(lr){18-19}
        \cmidrule(lr){20-21}

        % ==========================================
        % Second header row: Translation / Rotation
        % ==========================================

        &
        \scriptsize Trans. & \scriptsize Rot.
        &
        \scriptsize Trans. & \scriptsize Rot.
        &
        \scriptsize Trans. & \scriptsize Rot.
        &
        \scriptsize Trans. & \scriptsize Rot.
        &
        \scriptsize Trans. & \scriptsize Rot.
        &
        \scriptsize Trans. & \scriptsize Rot.
        &
        \scriptsize Trans. & \scriptsize Rot.
        &
        \scriptsize Trans. & \scriptsize Rot.
        &
        \scriptsize Trans. & \scriptsize Rot.
        &
        \scriptsize Trans. & \scriptsize Rot.
        \\

        \midrule

        % ==========================================
        % From Scratch
        % ==========================================

        From Scratch
        & 10.18 & 10.01
        & 9.62 & 10.20
        & 10.15 & 10.67
        & 9.79 & 11.08
        & 9.44 & 10.96
        & 10.75 & 11.43
        & 10.63 & 10.34
        & 11.00 & 12.04
        & 10.81 & 12.32
        & 11.03 & 14.09
        \\

        % ==========================================
        % EchoCLIP
        % ==========================================

        EchoCLIP~\cite{christensen2024vision}
        & \secondbest{8.10} & 7.33
        & \secondbest{7.69} & \secondbest{7.65}
        & 8.24 & \secondbest{8.11}
        & \secondbest{7.55} & \secondbest{8.37}
        & \secondbest{7.54} & \secondbest{8.28}
        & \secondbest{8.94} & \secondbest{8.98}
        & 8.42 & 7.40
        & 8.56 & 9.70
        & 8.69 & \secondbest{9.58}
        & 8.55 & 11.51
        \\

        % ==========================================
        % UltraSeP
        % ==========================================

        UltraSeP~\cite{jiang2025ultrasep}
        & 8.22 & \secondbest{7.25}
        & 7.77 & 7.87
        & \secondbest{8.18} & 8.20
        & 7.62 & 8.57
        & 7.55 & 8.44
        & 9.06 & 9.08
        & \secondbest{8.16} & \secondbest{7.31}
        & \secondbest{8.34} & \secondbest{9.44}
        & \secondbest{8.49} & 9.71
        & \secondbest{8.39} & \secondbest{11.30}
        \\

        % ==========================================
        % EchoACE (Ours)
        % ==========================================

        \textbf{EchoACE (Ours)}
        & \best{7.66} & \best{6.82}
        & \best{7.41} & \best{7.24}
        & \best{7.97} & \best{7.51}
        & \best{7.17} & \best{7.98}
        & \best{7.16} & \best{7.80}
        & \best{8.42} & \best{8.46}
        & \best{7.99} & \best{6.90}
        & \best{8.11} & \best{9.16}
        & \best{8.31} & \best{8.84}
        & \best{8.15} & \best{10.57}
        \\

        \bottomrule

    \end{tabularx}

    \endgroup

    \vspace{-5pt}

\end{table*}

% Category column
\newcolumntype{Q}{
    >{\raggedright\arraybackslash}m{1.55cm}
}

% Method column
\newcolumntype{R}{
    >{\raggedright\arraybackslash}m{2.05cm}
}

% Centered expandable numerical columns
\newcolumntype{Y}{
    >{\centering\arraybackslash}X
}

% ============================================================
% Table
% ============================================================

\begin{table*}[t]
    \centering

    \caption{
        \textbf{Per-view errors} of the \textbf{single-frame Action Model} across the \textbf{ablation} settings reported in the main paper's \textbf{Table~\ref{tab:loss_ablation}} across ten standard
        views.
        Translation and rotation errors (Trans./Rot.) are reported
        as mean absolute errors in millimeters and degrees,
        respectively.
    }

    \label{tab:ablation_per_view}

    \vspace{-2mm}

    \begingroup

    \footnotesize

    % Horizontal spacing
    \setlength{\tabcolsep}{0.4pt}

    % Vertical spacing
    \renewcommand{\arraystretch}{1.05}

    % Vertically center all X columns
    \renewcommand{\tabularxcolumn}[1]{m{#1}}

    % ========================================================
    % 22 columns:
    % 1: Category
    % 2: Method
    % 3-22: 10 views, each with Trans. and Rot.
    %
    % Vertical separator after Category column.
    % ========================================================

    \begin{tabularx}{\textwidth}{
        @{}
        Q
        !{\vrule width 0.4pt}
        R
        *{20}{Y}
        @{}
    }

        \toprule

        % ==========================================
        % First header row: Echocardiographic views
        % ==========================================

        \multicolumn{2}{c}{} 
        &
        \multicolumn{2}{c}{
            \scriptsize\bfseries PLAX
        }
        &
        \multicolumn{2}{c}{
            \scriptsize\bfseries PSAX-AV
        }
        &
        \multicolumn{2}{c}{
            \scriptsize\bfseries PSAX-PV
        }
        &
        \multicolumn{2}{c}{
            \scriptsize\bfseries PSAX-MV
        }
        &
        \multicolumn{2}{c}{
            \scriptsize\bfseries PSAX-PAP
        }
        &
        \multicolumn{2}{c}{
            \scriptsize\bfseries PSAX-APEX
        }
        &
        \multicolumn{2}{c}{
            \scriptsize\bfseries A4C
        }
        &
        \multicolumn{2}{c}{
            \scriptsize\bfseries A5C
        }
        &
        \multicolumn{2}{c}{
            \scriptsize\bfseries A3C
        }
        &
        \multicolumn{2}{c}{
            \scriptsize\bfseries A2C
        }
        \\

        \cmidrule(lr){3-4}
        \cmidrule(lr){5-6}
        \cmidrule(lr){7-8}
        \cmidrule(lr){9-10}
        \cmidrule(lr){11-12}
        \cmidrule(lr){13-14}
        \cmidrule(lr){15-16}
        \cmidrule(lr){17-18}
        \cmidrule(lr){19-20}
        \cmidrule(lr){21-22}

        % ==========================================
        % Second header row: Translation / Rotation
        % ==========================================

        \multicolumn{2}{c}{}
        &
        \scriptsize Trans. & \scriptsize Rot.
        &
        \scriptsize Trans. & \scriptsize Rot.
        &
        \scriptsize Trans. & \scriptsize Rot.
        &
        \scriptsize Trans. & \scriptsize Rot.
        &
        \scriptsize Trans. & \scriptsize Rot.
        &
        \scriptsize Trans. & \scriptsize Rot.
        &
        \scriptsize Trans. & \scriptsize Rot.
        &
        \scriptsize Trans. & \scriptsize Rot.
        &
        \scriptsize Trans. & \scriptsize Rot.
        &
        \scriptsize Trans. & \scriptsize Rot.
        \\

        \midrule

        % ====================================================
        % (1) Ablation study of different loss formulations
        % ====================================================

        \multicolumn{22}{l}{
            \footnotesize\bfseries\itshape
            (1) Ablation study of different loss formulations in EchoACE
        }
        \\

        \midrule

        % ------------------------------
        % Regression
        % ------------------------------

        \multirow[c]{2}{*}{Regression}
        &
        Direct Reg.
        & 17.64 & 14.34
        & 17.13 & 14.24
        & 18.06 & 15.24
        & 16.87 & 14.87
        & 16.17 & 14.87
        & 15.99 & 15.69
        & 18.27 & 16.70
        & 18.89 & 17.74
        & 18.67 & 16.42
        & 18.46 & 18.22
        \\

        &
        Scaled Reg.
        & 17.35 & \secondbest{13.57}
        & 16.21 & 13.38
        & 17.56 & 14.25
        & 16.28 & \secondbest{14.26}
        & 15.43 & \secondbest{14.32}
        & 15.22 & \secondbest{14.94}
        & \secondbest{16.84} & \secondbest{14.96}
        & \secondbest{17.43} & \secondbest{16.72}
        & 17.28 & \secondbest{15.68}
        & 17.24 & \secondbest{17.67}
        \\

        \midrule

        % ------------------------------
        % Ranking
        % ------------------------------

        \multirow[c]{2}{*}{Ranking}

        &
        ListNet
        & \secondbest{15.25} & 14.55
        & \secondbest{12.77} & \secondbest{11.83}
        & \secondbest{14.84} & \secondbest{13.22}
        & \secondbest{13.66} & 15.31
        & \secondbest{13.12} & 15.72
        & \secondbest{13.89} & 15.99
        & 17.46 & 15.17
        & 18.71 & 17.13
        & \secondbest{17.18} & 17.94
        & \secondbest{15.59} & 18.27
        \\
        
        &
        \textbf{RankNet (Ours)}
        & \best{10.24} & \best{8.64}
        & \best{9.10} & \best{8.66}
        & \best{9.73} & \best{9.01}
        & \best{9.29} & \best{9.45}
        & \best{8.88} & \best{9.62}
        & \best{10.04} & \best{9.99}
        & \best{10.44} & \best{8.66}
        & \best{11.14} & \best{10.46}
        & \best{10.21} & \best{11.31}
        & \best{10.42} & \best{12.59}
        \\

        % Double horizontal rule between the two studies
        \midrule
        \midrule

        % ====================================================
        % (2) Ablation study of Stage 1 and Stage 2 Pretraining
        % ====================================================

        \multicolumn{22}{l}{
            \footnotesize\bfseries\itshape
            (2) Ablation study of the contributions of Stage 1 and Stage 2 Pretraining in EchoACE
        }
        \\

        \midrule

        % ------------------------------
        % Stage 1 only
        % ------------------------------

        \multicolumn{2}{l}{
            Stage1-only Pretrain
        }
        & 16.66 & 19.26
        & 15.15 & 17.42
        & 16.48 & 18.12
        & 15.32 & 18.96
        & 14.79 & 19.22
        & 15.76 & 19.65
        & 19.40 & 19.31
        & 20.23 & 20.92
        & 20.83 & 22.10
        & 19.86 & 21.50
        \\

        % ------------------------------
        % Stage 2 only
        % ------------------------------

        \multicolumn{2}{l}{
            Stage2-only Pretrain
        }
        & \secondbest{13.38} & \secondbest{12.09}
        & \secondbest{11.78} & \secondbest{11.09}
        & \secondbest{13.08} & \secondbest{11.94}
        & \secondbest{12.24} & \secondbest{12.66}
        & \secondbest{11.91} & \secondbest{12.73}
        & \secondbest{12.79} & \secondbest{13.11}
        & \secondbest{14.34} & \secondbest{13.14}
        & \secondbest{15.12} & \secondbest{14.64}
        & \secondbest{14.28} & \secondbest{14.07}
        & \secondbest{14.51} & \secondbest{15.30}
        \\

        % ------------------------------
        % Stage 1 + Stage 2
        % ------------------------------

        \multicolumn{2}{l}{
            \textbf{Stage1+Stage2 Pretrain (Ours)}
        }
        & \best{10.24} & \best{8.64}
        & \best{9.10} & \best{8.66}
        & \best{9.73} & \best{9.01}
        & \best{9.29} & \best{9.45}
        & \best{8.88} & \best{9.62}
        & \best{10.04} & \best{9.99}
        & \best{10.44} & \best{8.66}
        & \best{11.14} & \best{10.46}
        & \best{10.21} & \best{11.31}
        & \best{10.42} & \best{12.59}
        \\

        \bottomrule

    \end{tabularx}

    \endgroup

    \vspace{-5pt}

\end{table*}

\end{document}